\documentclass{article}

\usepackage[final]{neurips_2024}

\usepackage[utf8]{inputenc} 
\usepackage[T1]{fontenc}    
\usepackage{hyperref}       
\usepackage{url}            
\usepackage{booktabs}       
\usepackage{amsfonts}       
\usepackage{nicefrac}       
\usepackage{microtype}      
\usepackage{xcolor}         

\usepackage{algorithm}
\usepackage{algorithmicx}
\usepackage{algpseudocode}

\usepackage{graphicx}
\usepackage{subfigure}
\usepackage{amsmath}
\usepackage{color}
\usepackage{bbm}
\usepackage{dsfont}
\usepackage{amsfonts}
\usepackage{pifont}
\usepackage{wrapfig}
\usepackage{caption}
\usepackage{comment} 
\usepackage{enumitem}
\usepackage{authblk}

\hypersetup{hypertex=true,
colorlinks=true,
linkcolor=red,
anchorcolor=green,
citecolor=green}
\usepackage{multirow} 

\newcommand{\ie}{\textit{i}.\textit{e}.}
\newcommand{\eg}{\textit{e}.\textit{g}.}

\newtheorem{theorem}{Theorem}[section] 

\newtheorem{observation}[theorem]{Observation}

\title{UMFC: Unsupervised Multi-Domain Feature Calibration for Vision-Language Models }

\author{%
  Jiachen Liang$^{1,2}$,
  Ruibing Hou$^{1}$\thanks{Corresponding author}
  , Minyang Hu$^{1,2}$,
  Hong Chang$^{1,2}$,
  Shiguang Shan$^{1,2}$,
  Xilin Chen$^{1,2}$\\
  $^1$  Institute of Computing Technology, Chinese Academy of Sciences\\
  $^2$University of Chinese Academy of Sciences\\
  \{jiachen.liang, minyang.hu\}@vipl.ict.ac.cn,
  \{houruibing, changhong, sgshan, xlchen\}@ict.ac.cn
}
\begin{document}

\maketitle

\begin{abstract}
Pre-trained vision-language models (\eg, CLIP) have shown powerful zero-shot transfer capabilities. But they still struggle with domain shifts and typically require labeled data to adapt to downstream tasks, which could be costly. In this work, we aim to  leverage unlabeled data that naturally spans multiple domains to enhance the transferability of vision-language models. Under this unsupervised multi-domain setting, we have identified inherent model bias within CLIP, notably in its visual and text encoders. Specifically, we observe that CLIP’s visual encoder tends to prioritize  encoding domain over discriminative category information, meanwhile its text encoder exhibits a preference for domain-relevant classes. To mitigate this model bias, we propose a \textit{training-free} and \textit{label-free} feature calibration method, Unsupervised Multi-domain Feature Calibration (UMFC). UMFC estimates image-level biases from domain-specific features and text-level biases from the direction of domain transition. These biases are subsequently   subtracted from original image and text features separately, to render them domain-invariant. We evaluate our method on multiple settings including transductive learning and test-time adaptation. Extensive experiments show that our method outperforms CLIP and performs on par with the state-of-the-arts that need additional annotations or optimization.
Our code is available at \url{https://github.com/GIT-LJc/UMFC}.

\end{abstract}

\section{Introduction}

Recently, Vision-Language Foundation Models (VLFMs) such as CLIP \cite{radford2021clip}, BLIP \cite{li2022blip}, Flamingo \cite{alayrac2022flamingo} and ALIGN \cite{jia2021align} have demonstrated remarkable performance across various downstream tasks. These VLFMs formulate the training objective as contrastive learning, leveraging millions of image-text pairs to establish a shared embedding space. Equipped with a wide range of visual and text representations, VLFMs exhibit the capability to tackle downstream tasks in a zero-shot manner.

Despite VLFMs being exposed to abundant examples, they may still encounter examples with new variations in downstream tasks.
To address the problem of distribution shift between the pre-training and downstream domains, a natural approach involves fine-tuning VLFMs on various target tasks, 
such as prompt engineering \cite{zhou2022coop, zhou2022cocoop} and adapter learning \cite{gao2024clipadapter, Zhang2022TipAdapter}. 
However, these methods generally require labeled samples for fine-tuning, which is prohibitively expensive to be satisfied in reality.
Conversely, abundant unlabeled data are often available for downstream tasks. Notably, in practical scenarios, the unlabeled data typically contains multiple domains, which exacerbates the adaptation difficulty of VLFMs. Therefore, in this paper, we aim to improve the adaptation performance of VLFMs directly on unlabeled data that naturally spans multiple domains.

\begin{figure}[t]
\captionsetup{font={small}}
    \centering   
    \subfigure[] 
    {
        \begin{minipage}[b]{.3\linewidth} 
            \centering
            \includegraphics[scale=0.2]{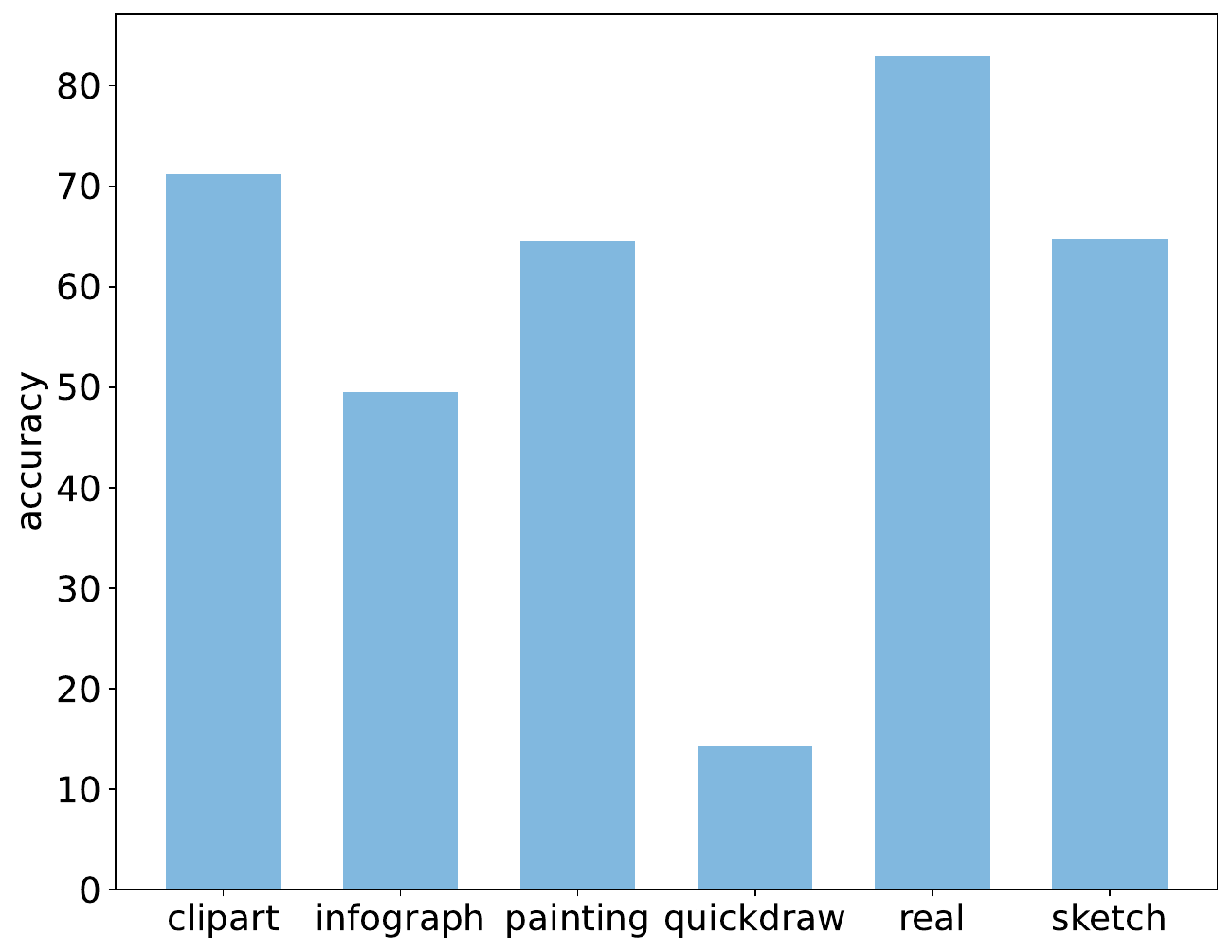}
        \end{minipage}
        \label{0_shot_accs}
    }
    \subfigure[]
    {
        \begin{minipage}[b]{.3\linewidth}
            \centering
            \includegraphics[scale=0.2]{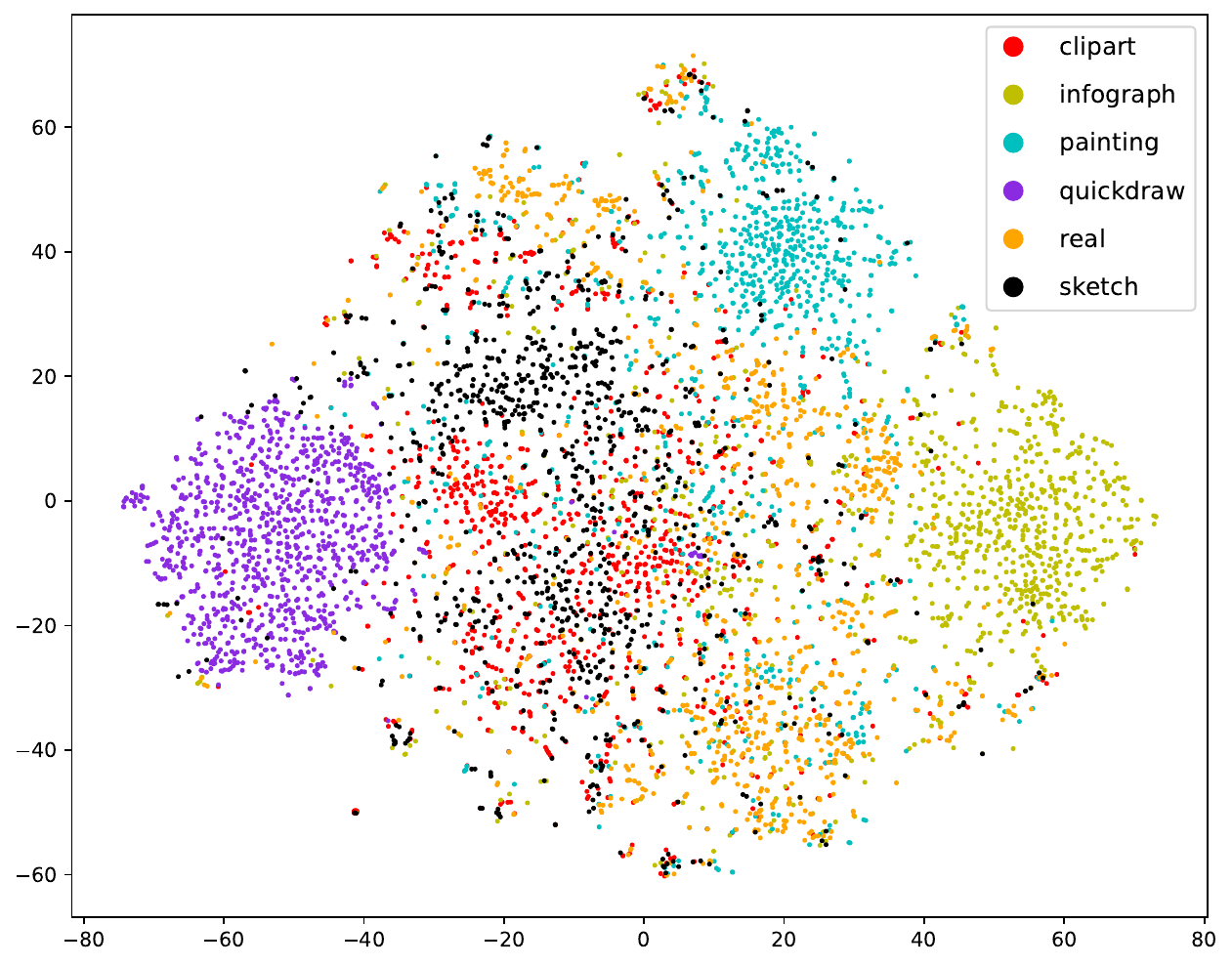}
        \end{minipage}
        \label{image_domain_clip}
    }
    \subfigure[]
    {
        \begin{minipage}[b]{.3\linewidth}
            \centering
            \includegraphics[scale=0.2]{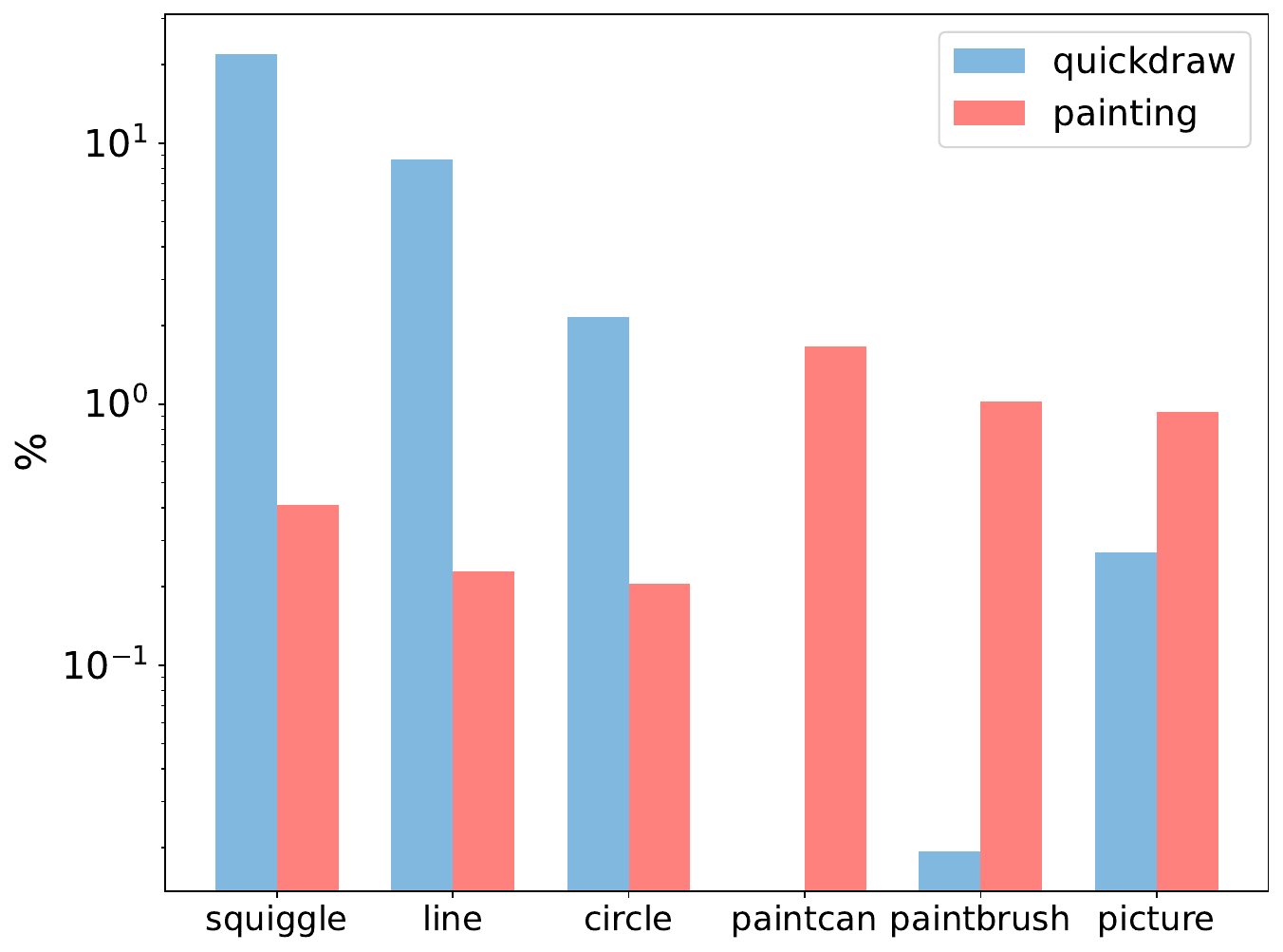}
        \end{minipage}
        \label{preds_numbers}
    }
    \caption{On DomainNet dataset, we visualize (a) The accuracy of CLIP on the six domains. (b) The image features extracted by CLIP's image encoder across different domains. The visualization show that CLIP exhibits inherent model bias. (c) The number of predictions for different classes on \textit{quickdraw} and \textit{painting} domains.}
    \label{introduction}
    \vspace{-0.8cm}
\end{figure}

In this unsupervised multi-domain setting, we observe that CLIP cannot perform well when unlabeled data are drawn from mixed distributions. 
As shown in 
Figure \ref{0_shot_accs}, even within the same class space, the accuracy of CLIP varies significantly across different domains.
While CLIP performs exceptionally well for images from common distributions encountered during pre-training, \eg, achieving   $83.0\%$ accuracy on \textit{real} domain, it struggles with rare distributions encountered during pre-training, \eg, only achieving $14.2\%$ accuracy on   \textit{quickdraw} domain.
Above observations highlight that CLIP exhibits \textit{model biases} that lead to incorrect predictions in specific scenarios.
This raises a fundamental question: \textit{where do these biases originate}?

We point out that these model biases stem from deficiencies in the visual encoder and textual encoder. 
\textit{On the side of visual encoder, we observe that CLIP’ visual encoder prioritizes encoding domain information over discriminative category information.} 
As shown in Figure \ref{image_domain_clip}, features from the same domain clearly cluster together, whereas a notable gap  separates features from different domains.
This phenomenon indicates that CLIP's visual encoder exhibits a higher sensitivity to domain information over category information. 
When the domain of downstream tasks shifts from pre-trained tasks, the mismatch in domain-specific information encoded in image features could adversely affect classification accuracy.  
\textit{On the side of textual encoder, we observe that CLIP demonstrates varying category preferences across different domains.} Specifically, CLIP tends to classify images into categories  whose name are closely related to corresponding domain.
As shown in Figure \ref{preds_numbers}, within ``quickdraw'' domain, a large portion of samples ($\sim 30\%$) are classified as ``squiggle'' or ``line'' categories. Conversely, within ``painting'' domain, CLIP favors categories like ``paintcan'' and ``paintbrush''. This observation shows that the class embeddings encoded by CLIP’s textual encoder  inherently carry  domain-specific information, misleading the model to prioritize categories highly associated with respective domains. Due to the combined effects of visual and textual encoder biases, CLIP's performances vary significantly across different domains. 

In this paper, we aim to calibrate CLIP to mitigate the model biases. Initially, we analyze the model biases from a probabilistic standpoint. 
The factors influencing $p(y|x)$, affected by domain variable $z$, can be decoupled into two parts: the sample distribution conditioned on classes $p(x|y, z)$ and the class distribution $p(y|z)$. If the two probability distributions are independent of $z$, domain shifts will not affect the predictions.
To this end, we propose \textbf{Unsupervised Multi-domain Feature Calibration} (UMFC), a simple yet efficient framework for calibrating CLIP to generalize to various downstream tasks using multi-domain unlabeled data. UMFC jointly calibrates CLIP through two \textit{training-free} modules: Image Feature Calibration module (IFC) and Text Feature Calibration module (TFC). 
\textbf{Firstly}, IFC focuses on calibrating CLIP's image encoder to prioritize category-level over domain-level information, thereby reducing  prediction error caused by domain shifts. Specifically, we calculate the average image features for each domain $i$, denoted as $\mu_i$, 
and posit that the prediction of $\mu_i$ reflects the inherent bias of CLIP's image encoder within that domain. By subtracting this domain-specific bias from original predictions,  we can derive domain-agnostic predictions. 
\textbf{Secondly}, TFC focuses on calibrating CLIP's text encoder to remove its preference towards domain-related class names. 
As observed by \cite{Dunlap2022lads, Patashnik2021StyleCLIP, fahes2023poda}, a ``global direction'' exists in CLIP, representing the shift from training distribution to unseen distribution, shared across image and text embedding spaces. Motivated by this observation, we utilize the shift direction between different-domain images to estimate the text shift direction.
Then, we subtract this shift vector to counteract the text encoder's bias towards domain-related categories. By combing IFC and TFC, we can calibrate the features on both CLIP's image and text encoders, thereby alleviating the model bias in downstream tasks.

We validate the efficacy of UMFC on three downstream tasks: unsupervised calibration, transductive learning, and test-time adaptation, demonstrating consistent gains over CLIP. 
UMFC presents a low-cost solution for classification,  unlocking the potential of CLIP-like models for practical scenarios characterized by abundant images across multiple domains but scarce labels.

\section{Related Work}

\textbf{Few-Shot Learning.} Few-shot learning (FSL) \cite{hu2023tad, hu2024tadv2, hu2024smn, finn2017maml, snell2017prototypical} aims to learn a good model on a novel task with few labeled samples. Traditional FSL methods \cite{hu2023tad, hu2024tadv2, hu2024smn, finn2017maml, snell2017prototypical}  often rely on abundant related training tasks for pre-training and design specialized algorithms to transfer cross-task knowledge, facilitating task adaptation.
Additionally, some semi-supervised learning (SSL) methods \cite{liang2023ssfa, hou2023dual, hou2024triplet} tackle the scarcity of labeled samples by assuming access to extensive unlabeled data. 
While both FSL and SSL methods have achieved encouraging results, their generalization abilities are limited. 
Recently, the CLIP model is proposed, which is a vision-language foundation model that learns a shared vision-language embedding space. 
As pre-trained on large-scale data, CLIP exhibits impressive zero-shot ability across various downstream tasks.
Based on CLIP model, many works focus on improving its performance further on downstream tasks with few labeled data.
For example, 
CoOp \cite{zhou2022coop}, CoCoOp \cite{zhou2022cocoop}, MaPLe \cite{khattakMaPLe} and PromptSRC \cite{Khattak_2023_promptsrc} leverage few labeled samples to learn the prompt in the continual input embedding space, offering a parameter-efficient way of fine-tuning foundation models. 
Similarly, 
CLIP-Adapter \cite{gao2024clipadapter} and Tip-Adapter-F \cite{Zhang2022TipAdapter} introduce a lightweight adapter module to produce adaptive multi-modal features.  
However, these methods require extra labeled data and the process of tuning pre-trained parameters for CLIP, which is cost-expensive.
Differently, in this work, we focus on utilizing unlabeled data to enhance CLIP's performance in a training-free manner.


\textbf{Domain Adaptation / Domain Generalization.} \
Recently, several methods \cite{Lai2023PADCLIP, Chen2022MPA, fahes2023poda, Ruan2021CAD, Cha2022MIRO, Shu2023CLIPood, Cho2023PromptStyler, Huang2023RISE, Dunlap2022lads} exploit a large-scale pre-trained model (\eg, CLIP) to address the domain adaptation and generalization problems.
For example, RISE \cite{Huang2023RISE} leverages CLIP as a teacher to regularize the student’s learned representation through images. The work \cite{Chen2022MPA} utilizes domain-invariant and domain-specific prompts for multi-source unsupervised domain adaptation. 
Other works \cite{fahes2023poda, Dunlap2022lads, Cho2023PromptStyler} focus on utilizing the transferability between visual and textual modalities to guide domain information transfer. 
PromptStyler \cite{Cho2023PromptStyler} attempts to simulate various distribution shifts 
to explore diverse styles in a joint vision-language space. PODA \cite{fahes2023poda} and LADS \cite{Dunlap2022lads} generate samples in the style of the target domain, and adapt to the target domain based on these samples.
However, these language-guided methods require the prior of target-domain names, which may not be satisfied in reality. In contrast, our UMFC method does not require any extra information about the target domain.

\textbf{Test-Time Adaptation.} \
Test-time adaptation aims to adapt a pre-trained model to test tasks, where distribution of test data differs from that of pre-training data.
TPT \cite{Shu2022TPT} proposes a test-time prompt tuning strategy, 
which extends traditional TTA methods to vision-language models. 
Building upon TPT, DiffTPT \cite{Feng2023DiffusionTPT} utilizes pre-trained diffusion models to augment the diversity of test data samples used in prompt tuning.
SwapPrompt \cite{Ma2023SwapPrompt} employs a dual prompts paradigm to enhance the swapped prediction mechanism. 
However, these prompt learning methods are computationally expensive and time-consuming.
Different from above methods, our UMFC only needs to calibrate features 
in a training-free way, making it more efficient for test-time adaptation.

\section{CLIP and Model Biases}
In this section, we first describe  the backgrounds of CLIP and then analyze its bias issue in downstream tasks. We attribute the cause of bias to the visual encoder bias and text encoder bias.

\subsection{Contrastive Language-Image Pre-training (CLIP)}
CLIP \cite{radford2021clip} 
consists of two parallel encoders: 
a visual encoder that maps image inputs into image features, and a text encoder that maps text inputs into text features. The model is trained with a contrastive loss that maximizes similarity between positive image-text pairs while minimizing similarity between negative pairs. This process ensures alignment between the features of images and their corresponding textual descriptions within the feature space. Trained on a vast collection of image-text pairs,  CLIP benefits from general visual representations, endowing it with powerful zero-shot capabilities.

Formally, we denote a CLIP model as $\left\{F, T\right\}$, with $F$ and $T$ being the visual and text encoders. Considering a $K$-class classification problem, we  use $\mathcal{Y}=\left\{y_1, \dots, y_K \right\}$ to represent the class space, where $y_k$ denotes the class name of $k^{th}$ class. In the zero-shot inference phase, CLIP uses hand-crafted prompts $\mathrm{p}$ (such as ``a photo of a \{\}'') to covert each class name $y_i$ to category-specific text  description $\left\{\mathrm{p};y_i\right\}$. Then, we  feed these class descriptions to the text encoder to get the text features $\left\{t_1, \dots, t_K\right\}$, where $t_i=T\left(\mathrm{p};y_i\right)$. 
Meanwhile, given a test image $x$, the visual encoder is used to compute its visual feature, denoted as $f=F\left(x\right)$. The prediction probability on $x$ can be then computed as: 
\begin{equation}
    p(y_i|x)=\frac{\exp({\mathrm{sim}(f, t_i)/ \tau})}{\sum_{j=1}^K{\exp({\mathrm{sim}(f, t_j)/ \tau})}},
    \label{eq:conditional prob}
\end{equation}
where $\mathrm{sim}$ denotes the cosine similarity and $\tau$ is the temperature of the softmax function.

\subsection{Model Bias in CLIP}
\label{sec:3.2}
As shown in Figure \ref{0_shot_accs}, the accuracy of CLIP varies significantly across different domains within the same class space, \eg, $83.0\%$ in \textit{real} domain and $14.2\%$ in \textit{quickdraw} domain. This phenomenon indicates that CLIP favors domains commonly encountered during pre-training (such as natural images). We attribute this model bias to visual encoder bias and text encoder bias. 
\begin{itemize}[leftmargin=*]
\item \textbf{Visual Encoder Bias.}  \ We observe that CLIP’ visual encoder prioritizes encoding domain information over discriminative category information, as shown in Figure \ref{image_domain_clip}. Due to the abundance of natural images in pre-training, CLIP's visual encoder and text encoder are well-aligned in the natural image domain. Consequently, when the style of input images largely deviates from natural domains (such as \textit{quickdraw} and \textit{infograph} style), the visual feature gap across different domain (as depicted in Figure \ref{image_domain_clip}) hinders the text encoder's capacity to  process these shifted images effectively, leading a significant drop in classification accuracy.
\item  \textbf{Text Encoder Bias.} \ We observe that CLIP exhibits a preference for domain-related categories in specific  domains, as shown in Figure \ref{preds_numbers}. 
In domains characterized by distinct styles, certain category names may carry domain-specific information. For instance, in \textit{quickdraw} domain, where most images consist of lines and squiggle, CLIP demonstrates a severe bias towards the ``line'' and ``squiggle'' categories. This leads to a large number of quickdraw samples being incorrectly classified into the two categories.  Conversely, in the \textit{painting} domain, as all images inherently possess painting features, CLIP shows a strong preference for categories related to the concept of \textit{painting}, such as ``paintcan'' and ``paintbrush''.
\end{itemize}

\section{Unsupervised Multi-domain Feature Calibration}

In this section, we continue to analyze the model biases issue from a probabilistic view. 
Then we give a detailed introduction of our method aimed at alleviating CLIP's biases.

\subsection{Analyze Model Biases in a Probabilistic View}
We start from the posterior probability $p\left(y_i|x\right)$, as introduced in Equation \ref{eq:conditional prob}.
Recall that our goal is to maximize the posterior probability $p\left(y_i|x\right)$ for each image-label pair $\left(x, y_i\right)$ of any domain.
A natural question arises: \emph{How different domains affect the posterior probability?}
We answer this question from a probabilistic view.
Based on the Bayes' Theorem, we can drive that:
\begin{align}
    p\left(y_i|x\right) &= \frac{p\left(y_i, x\right)}{p\left(x\right)} = \frac{\sum_{z} p\left(y_i, x, z\right)}{p\left(x\right)},
    \label{eq:bayes equation}
\end{align}
where $z$ is a latent variable that denotes the domain labels.
We can further decompose the joint probability $p\left(y_i, x, z\right)$ in Equation \ref{eq:bayes equation} as $p\left(y_i, x, z\right) = p\left(x|y_i, z\right)p\left(y_i|z\right)p\left(z\right)$.
Assume that the probability distribution of domains $p\left(z\right)$ is uniform, thus maximizing the posterior probability $p\left(y_i|x\right)$ is equal to maximize the summation $\sum_{z} p\left(x|y_i, z\right)p\left(y_i|z\right)$, as
\begin{align}
    \max p\left(y_i|x\right) &= \max \frac{\sum_{z} p\left(x|y_i, z\right)p\left(y_i|z\right)p\left(z\right)}{p\left(x\right)} \notag\\
    &= \max \sum_{z} p\left(x|y_i, z\right)p\left(y_i|z\right).
    \label{eq3}
\end{align}
Equation \ref{eq3} shows that the domains affect posterior probability distribution $p\left(y|x\right)$ by disturbing two terms: the sample distribution conditioned on classes $p\left(x|y,z\right)$ and the class distribution $p\left(y|z\right)$.

\subsection{UMFC: Unsupervised Multi-domain Feature Calibration}

As analyzed in Section \ref{sec:3.2},  CLIP exhibits both visual and text encoder bias, impacting its generalization ability to downstream tasks.
As shown in Equation \ref{eq3}, these biases stem from the probability distributions $p\left(x|y,z\right)$ and $p\left(y|z\right)$, which are disturbed by domain information.
To mitigate the model biases, an intuitive idea is to make the two probability distributions $p\left(x|y\right)$ and $p\left(y\right)$ independent of domain variable $z$ by calibrating image and text features.  
To this end, we propose \textit{training-free} UMFC approach, consisting of an Image Feature Calibration (IFC) module to alleviate visual encoder bias and Text Feature Calibration (TFC) module to alleviate text encoder bias, facilitating the transfer of CLIP to downstream tasks. Algorithms are provided in the Appendix \ref{appendix: algorithm}.

\textbf{Image Feature Calibration Module.} \
On the side of visual encoder, we focus on making the conditional probability $p\left(x|y\right)$ independent of domain variable $z$, which is equivalent to achieving $p\left(x|y, z\right) = p\left(x|y\right)$.
A straightforward method is to align image feature distributions given a class $y$ across different domains.
Unfortunately, as only mixed unlabeled data is provided, we cannot access to class labels and domain labels of images.
However, we empirically observe that CLIP's visual encoder prioritizes encoding domain information over discriminative category information.
Thus, we can distinguish image features from different domains through a simple clustering algorithm.
Formally, we assume that there are $M$ clusters $\{c_1,...,c_M\}$ after applying a clustering algorithm. 
Each cluster is assumed to follow a Gaussian distribution $c_i \sim \mathcal{N}(\mu_i, \Sigma_i)$, where mean vector $\mu_i = (\frac{1}{|c_i|}\sum_{f\in c_i} f)$ represents the average of image features from cluster $c_i$.
Due to the model biases, the pseudo labels produced by zero-shot CLIP are not reliable.
Therefore, we directly align the margin image feature distribution of each cluster by subtracting the corresponding mean vector\footnote{This operation is based on the assumption of a uniform class distribution in each cluster. 
We experimentally found that this operation remains effective, even if this assumption does not hold strictly.}.
Specifically, for each visual feature $f$ belonging to the cluster $c_i$, we calibrate it as follows:
\begin{equation}
    f' = \frac{f - \mu_i}{\left\|f - \mu_i\right\|_2}.
    \label{eq_IFC}
\end{equation}

\textbf{Text Feature Calibration Module.} \
On the side of text encoder, we focus on making the class  probability $p\left(y\right)$ independent of domain variable $z$, which is equivalent to achieving $p\left(y|z\right) = p\left(y\right)$.
Previous works \cite{Wang2022DeFo} have found that CLIP's performance is sensitive to class name $y$.
For example, replacing category names with synonyms or near-synonymous terms can significantly impact CLIP's prediction results. 
Due to the influence of domain information in image features, text encoder bias can cause CLIP to categorize domain-specific images into categories whose names are semantically similar to that domain. 
We further observe that such sensitivity to class names varies across different domains, as shown in Figure \ref{preds_numbers}.
This observation inspires us to calibrate text features by removing domain-specific information.
However, a challenge arises as domain labels are unavailable.
To address this issue, we attempt to extract domain information from unlabeled images, and then transfer this domain information to the text embedding space to estimate the text bias.
\begin{observation}
\label{global_direction}
\textbf{Cross-Modality Transition Direction}. The underlying assumption behind using images to simulate the corresponding shifts in texts is that the transition direction from domain $i$ to domain $j$ is consistent across both the image embedding and text embedding spaces \cite{Dunlap2022lads, Patashnik2021StyleCLIP, fahes2023poda}, which can be formulated as:
\begin{equation}
\frac{F\left(x^i\right) - F\left(x^j\right)}{\left\|F\left(x^i\right) - F\left(x^j\right)\right\|_2} \approx \frac{T\left(\mathrm{p}^i;y^i\right) - T\left(\mathrm{p}^j;y^j\right)}{\left\|T\left(\mathrm{p}^i;y^i\right) - T\left(\mathrm{p}^j;y^j\right)\right\|_2},
\label{eq5}
\end{equation}
where $\left(x^i,y^i\right)$ and $\left(x^j,y^j\right)$ represent training samples from domain $i$ and domain $j$ respectively, 
$p^i$ and $p^j$ denote the domain-specific text prompts for domain $i$ and $j$ respectively. 
For example, the text prompt of ``quickdraw'' domain can be ``a quickdraw image of a [class]''.
\end{observation}
Inspired by Observation \ref{global_direction}, we estimate the text-level domain transition direction using different-domain images. By clustering image features, we calculate the average feature $\mu_{i}$ of unlabeled images from each domain $i$, representing the domain-specific feature for that domain. Also, we calculate the average feature of all unlabeled images $\mu_{\mathrm{avg}}$, representing the domain-invariant feature since it encompasses various domain distributions.
Following Equation \ref{eq5}, the transition information $\widehat{t^i}$ of domain $i$ can be computed as $\widehat{t^i} = \mu_{i} - \mu_{\mathrm{avg}}$. 
By subtracting this domain transition vector, we suppress the preference for specific class names in the original text features. To ensure that the calibrated text features effectively eliminate biases towards a wide range of domains, we integrate the text features calibrated on each domain. Specifically, for the text feature $t_j$ of class $j$, we calibrate it as follows:
\begin{equation}
    t_j' = \frac{1}{M}\sum_{i=1}^{M}{\frac{{t_j - \widehat{t^i}}}{\left\|{t_j - \widehat{t^i}}\right\|_2}},
    \label{eq_TFC}
\end{equation}
where $M$ is the number of clusters.

\textbf{Inference.}
After calibration with IFC and TFC modules, we obtain the final prediction results. Specifically, this calibration modifies the prediction probability on test image $x$ as follows:
\begin{equation}
p\left(y_i|x\right)=\frac{\exp\left({\mathrm{sim}\left(f',t'_i\right)/ \tau}\right)}{\sum_{j=1}^K{\exp\left({\mathrm{sim}\left(f',t'_j\right)/ \tau}\right)}}.
\label{final_preds}
\end{equation}

\section{Experiments}

\subsection{Experimental Setting}
\textbf{Datasets.} \
Our UMFC is \textit{training free}, which only calibrates the image and text features using Equation \ref{eq_IFC} and \ref{eq_TFC}. To analyze model's generalization capability, we use two large-scale datasets for evaluation: 1) \textit{DomainNet}  \cite{Leventidis2021DomainNet} consists of 569,010 images with 345 categories from six domains: Clipart (C), Infograph (I), Painting (P), Quickdraw (Q), Real (R), Sketch (S). 2) \textit{ImageNet Variants} composed of several datasets shifted from ImageNet, including ImageNet-A (IN-A) \cite{Hendrycks2019imagenet-a}, ImageNet-R (IN-R) \cite{Hendrycks2020imagenet-r}, and ImageNet-Sketch (IN-S) \cite{Wang2019imagenet-s}.  We form the class space for \textit{ImageNet Variants} by taking the union of the class sets in IN-A and IN-R.
To ensure the reliability of the evaluation results, we randomly sample the test data to construct a balanced test set where both domain and category distributions are uniform.

\textbf{Evaluation Paradigms.} \
Our approach is a universal feature calibration technique that can be applied across multiple scenarios. In this work, we explore three settings: 
1) \textit{Unsupervised Calibration} (UC) where the unlabeled training set is provided for computing the calibration vectors; 
2) \textit{Transductive Learning} (TL) where the entire unlabeled test set is provided at once, without providing any training data; 
3) \textit{Test-Time Adaptation} (TTA) where the model can be adapted to test samples shifted from training distribution.  Different from TL, the test data usually arrives in batches in TTA setting. 
More details on the experimental setup, please refer to Appendix \ref{appendix:experimental-setting}.

\subsection{Main Results on Unsupervised Calibration.}

\textbf{Implementation Details.} \
We select CLIP \cite{radford2021clip} as our pre-trained vision-language model.
We use CLIP with 
ViT-B/16 \cite{Dosovitskiy2020vit-b16} as image encoder, and keep the original transformer as the text encoder. By default, a fixed prompt, ``a photo of a [class]'', is employed for all datasets. The images are resized to $224\times 224$.
The hyper-parameter $M$ (cluster number) is set to $6$ for DomainNet.
Remarkably, our method is training-free where both image encoder and text encoder remain frozen throughout the entire pipeline. All experiments are performed on a GeForce RTX 3090 Ti GPU.

\textbf{Baselines.} \
We compare our method with four groups of methods: 
(1) CLIP and its variants to show zero-shot predictions:  CLIP \cite{radford2021clip} that uses a fixed prompt "a photo of a [class]"; CLIP-E \cite{radford2021clip} that uses an ensemble of prompt templates.
(2) CLIP-D \cite{radford2021clip} that utilizes the domain information of test samples and designs domain-specific prompts.
(3) Few-shot learning methods:  CoOp \cite{zhou2022coop} that performs prompt tuning using labeled data of downstream tasks; 
CLIP-Adapter \cite{gao2024clipadapter} that trains an adapter using task-specific labeled data. 
(4) Unsupervised Fine-tuning method: MUST \cite{Li2022MUST} that fine-tunes the model using unlabeled multi-domain data.

\textbf{Analysis.} \
As shown in Table \ref{DN-on-UC-Multi}, UMFC achieves superior performance over CLIP and competitive performance with few-shot methods, CoOp \cite{zhou2022coop} and CLIP-Adapter \cite{gao2024clipadapter}. We can observe that: 
\textbf{(1)} 
CLIP-D incorporates domain information into text prompts, such as  ``a clipart image of a [class]' for test samples from ``clipart'' domain. However, creating domain-specific templates is challenging due to unknown and potentially mixed domain sources of test samples. 
So, we only use CLIP-D as an oracle result. 
As shown in Table \ref{DN-on-UC-Multi}, our method can achieve competitive performance with CLIP-D without prior domain labels for test samples.
\textbf{(2)} On domains where CLIP performs poorly, \eg, ``quickdraw'', our UMFC significantly outperforms others. 
Compared to CLIP, UMFC achieves about $5\%$ performance gain on quickdraw domain.
In addition, combined with more diverse prompts of CLIP-E, UMFC further improves performance.
\textbf{(3)} MUST \cite{Li2022MUST} uses abundant unlabeled data from 6 domains for fine-tuning. 
Our UMFC achieves better performance than MUST even without any additional training.
\textbf{(4)} 
For CoOp \cite{zhou2022coop} and CLIP-Adapter \cite{gao2024clipadapter}, we fine-tune  them  using labeled samples  from multiple domains. Specifically, we provide $k$ labeled samples per class for each domain, resulting in a total of $(6  \times k \times 345)$ labeled samples available for training. 
When the number of labeled samples is $6\times 345$, our method outperforms few-shot fine-tuning methods. While the performance is higher for few-shot methods with a larger number of labeled samples (\ie,  $24\times 345$),  it is essential to highlight the challenges in obtaining some labeled data for each class and each domain in real-world scenarios. 
 In contrast, our method does not require selecting class-balanced  and domain-balanced  labeled samples and incurs no additional training overhead.

Table \ref{DN-on-UC-Single} provides $8 \times 345$ labeled samples from a single domain for CoOp fine-tuning and an equal number of unlabeled samples for UMFC calibration. While CoOp trained on a single domain can improve performance within that domain, its performance declines on other domains. This decline becomes particularly notable when a significant distribution gap between the training and test domains exists, \eg, CoOp (Q), leading to a large decrease in average performance across multiple domains. 
In contrast, our method achieves consistent performance improvements on both training and unseen domains with the same amount of training data. Additionally, unlike CoOp, UMFC does not require labeled data or any parameter fine-tuning.

\begin{table}[]
\centering
\captionsetup{font={small}}
\scriptsize
\caption{Results on DomainNet under multi-domain Unsupervised Calibration. CLIP denotes zero-shot CLIP with a fixed text prompt template ``a photo of a [class]'', CLIP-E uses the ensemble prompt templates designed for Imagenet \cite{zhou2022coop}, CLIP-D uses the domain-specific templates, \ie, ``a [domain] image of [class]''. CoOp and CLIP-Adapter are trained on multi-domain labeled data, \eg, $6\times1\times345$ denotes the number of labeled data.}
\vspace{+2mm}
\label{DN-on-UC-Multi}
\begin{tabular}{l | l|cccccc|c}
\toprule
 & Method   & C& I                 & P                  & Q                 & R   & S & Avg\\
\midrule
 \multirow{6}{*}{Unsupervised} &    CLIP  \cite{radford2021clip}   & 71.21  & 49.47  & 64.61  & 14.23  & 82.98  & 64.81  & 57.88  \\
   & CLIP-E \cite{radford2021clip}            & {73.16} & {54.17} & {67.02} & {15.86} & {84.30} & {67.49} & {60.33} \\
   \cmidrule{2-9}
  &  CLIP-D   \cite{radford2021clip}      & {73.90} & {55.84} & {67.75} & {17.84} & {83.26} & {67.56} & {61.03} \\
   \cmidrule{2-9}
 &  MUST \cite{Li2022MUST}  & 74.83  & 56.48 & 61.80 & 19.06 & 82.88 & 70.31 & 60.89 \\
   \cmidrule{2-9}
  &   \textbf{UMFC} (ours) & 73.02 & 55.30 & 66.36 & 19.67 & 83.54 & 66.87 & 60.79 \\
 &   \textbf{UMFC + CLIP-E} (ours)  & 73.84 & 56.59 & 67.39 & 20.03 & 84.33 & 67.90 & \textbf{61.68} \\
    \cmidrule{1-9}
\multirow{4}{*}{Few-Shot}  &   CoOp ($6\times1\times345$) \cite{zhou2022coop}   & 72.73 & 53.95 & 66.80 & 19.58 & 82.53 & 67.27 & 60.48 \\
  &  CoOp   ($6\times4\times345$)  \cite{zhou2022coop}  & 74.7  & 54.96 & 68.29 & 22.14 & 82.94 & 69.48 & 62.09 \\
 &  CLIP-Adapter ($6\times1\times345$) \cite{gao2024clipadapter} & 72.67 & 51.69 & 67.84 & 17.82 & 84.26 & 65.75 & 60.00 \\
&    CLIP-Adapter ($6\times4\times345$) \cite{gao2024clipadapter} & 74.35 & 53.79 & 69.94 & 19.71 & 85.26 & 66.90 & 61.66 \\
   
\bottomrule
\end{tabular}

\end{table}

\begin{table}[]
\centering
\captionsetup{font={small}}
\scriptsize
\caption{Results on DomainNet under single-domain Unsupervised Calibration. $8 \times 345$ samples (each class has $8$ samples) from a single domain are provided. CoOp (C/Q/I) and UMFC (C/Q/I) denote training samples for CoOp and UMFC from the ``Clipart''/``Quickdraw''/``Infograph'' domains, respectively.}
\vspace{+2mm}
\label{DN-on-UC-Single}
\begin{tabular}{l|cccccc|c}
\toprule
Method   & C& I                 & P                  & Q                 & R   & S & Avg\\
\midrule
 
    CLIP   \cite{radford2021clip}    & 71.21  & 49.47  & 64.61  & 14.23  & 82.98  & 64.81  & 57.88  \\
    \cmidrule{1-8}
    CoOp (C) \cite{zhou2022coop}  & 74.55 & 42.66 & 55.94 & 13.82 & 75.00 & 58.73 & 53.45 \\
    \textbf{UMFC} (C)                      & 73.27 & 52.96 & 65.27 & 16.94 & 83.60 & 67.04 & \textbf{59.85} \\
    \cmidrule{1-8}
    CoOp (Q) \cite{zhou2022coop} & 43.97 & 25.5  & 32.63 & 29.07 & 48.44 & 38.74 & 36.39 \\
    \textbf{UMFC} (Q)                     & 72.17 & 49.65 & 63.85 & 17.47 & 82.84 & 66.36 & \textbf{58.72} \\
    \cmidrule{1-8}
    CoOp (I) \cite{zhou2022coop} & 60.19 & 54.28 & 50.81 & 11.19 & 70.73 & 54.27 & 50.24 \\
    \textbf{UMFC} (I)                     & 72.54 & 55.21 & 64.48 & 16.30 & 83.31 & 66.51 & \textbf{59.73} \\
   						
\bottomrule
\end{tabular}
\end{table}

\subsection{Main Results on Transductive Learning}

\begin{table}[t]
    \captionsetup{font={small}}
    \caption{Comparison  Results on DomainNet under Transductive Learning. }
    \vspace{+2mm}
    \centering
    \scriptsize
    
    \begin{tabular}{l|cccccc|c}
    \toprule
    Method    & C & I   & P  & Q    & R    & S  & Avg \\
    \midrule
    CLIP \cite{radford2021clip}     & 71.21     & 49.47   & 64.61    & 14.23   & 82.98   & 64.81   & 57.88    \\
    CLIP-E  \cite{radford2021clip}           & {73.16} & {54.17} & {67.02} & {15.86} & {84.30} & {67.49} & {60.33} \\
    CLIP-D \cite{radford2021clip}        & {73.90} & {55.84} & {67.75} & {17.84} & {83.26} & {67.56} & {61.03} \\
    MIRO \cite{Cha2022MIRO}             & {-} & {-} & {-} & {-}  & {-} & {-} & {54.00} \\
    PromptStyler \cite{Cho2023PromptStyler}            & {73.10} & {50.90} & {69.20} & {13.30}  & {85.40} & {65.30} & {59.40} \\
    \cmidrule{1-8}
    \textbf{UMFC}  & 73.01   & 55.44   & 66.89   & 20.14   & 83.66   & 67.51   & \textbf{61.11} \\  
    \bottomrule
    \end{tabular}
    \label{DN-on-TL}
\end{table}

In this part, we compare our UMFC with three groups of methods: 
(1) Zero-Shot CLIP models. 
(2) Domain Generalization (DG) method: MIRO \cite{Cha2022MIRO} that trains CLIP on available data to learn great generalization capability.
(3) Data-Free DG method: PromptStyler \cite{Cho2023PromptStyler} that learns to simulate diverse distributions. 
The hyper-parameter $M$ (cluster number) is set to $3$ for ImageNet-Variants.
The compared results are shown in Table \ref{DN-on-TL} and Table \ref{IN-on-TL}. As seen, our UMFC can achieve the best average performance on DomainNet and ImageNet-Variants benchmarks, which validates the effectiveness of our approach in transductive learning.

\begin{table}[H]
  \begin{minipage}[b]{0.48\linewidth}
    \centering
    \captionsetup{font={small}}
    \scriptsize
    
    \caption{Comparison Results on ImageNet-Variants under Transductive Learning.}
    \vspace{+2mm}
    \begin{tabular}{l|ccc|c}
    \toprule
    Method    & IN-A    & IN-R    & IN-S    & Avg \\
    \midrule
    CLIP   \cite{radford2021clip}   & 42.13   & 66.95   & 74.58   & 61.22   \\
    CLIP-E   \cite{radford2021clip}         & {45.42} & {71.10}  & {77.08} & {64.53} \\
    \cmidrule{1-5}
    \textbf{UMFC} & 43.42   & 68.86   & 77.24   & 63.17   \\
    \textbf{UMFC + CLIP-E} & {44.77}  & {72.19}  & {78.62}  & {\textbf{65.19}} \\  
    \bottomrule
    \end{tabular}
    
    \label{IN-on-TL}
  \end{minipage}
  \hspace{0.04\linewidth}
  \begin{minipage}[b]{0.48\linewidth}
    \centering
    \captionsetup{font={small}}
    \scriptsize
    
    \caption{Comparison Results on ImageNet-Variants under Test-Time Adaptation. }
    \vspace{+2mm}
    \label{IN-on-TTA}
    \begin{tabular}{l|ccc|c}
    \toprule
     Method              & IN-A   & IN-R   & IN-S   & Avg    \\
    \midrule
    CLIP                & 42.13  & 66.95  & 74.58  & 61.22  \\
    TPT \cite{Shu2022TPT}                & 47.16  & 59.95  & 67.49  
    & 58.20  \\
      \cmidrule{1-5}
    \textbf{UMFC-Memory} & 42.76  & 67.03  & 75.15  & 61.65  \\
    \textbf{UMFC-EMA}         & {42.21} & {67.82} & {76.26} & {\textbf{62.10}} \\
    \bottomrule
    \end{tabular}
  \end{minipage}
  \vspace{-0.25cm}
\end{table}

\subsection{Main Results on Test-Time Adaptation}
\textbf{Implementation Details.} \
In TTA setting, where we cannot access all data simultaneously, we adopt an incremental clustering approach. By default, the batch size is set to 100. 
 Initially, we apply K-Means clustering algorithm to the first batch of data. For subsequent batches, we use the prototype classification, with cluster centers serving as prototypes, to assign samples in that batch to different clusters. Then, the cluster centers and calibration statics $\left\{\mu_i\right\}_{i=1}^M$  are updated accordingly. 
 We have developed two strategies for updating the calibration statics:
 \textbf{UMFC-Memory} that stores the feature information of each batch to calculate the statistical information;
 \textbf{UMFC-EMA} that uses Exponential Moving Average (EMA) to update statistical information.

\begin{table}[H]
\centering
\captionsetup{font={small}}
\small
\scriptsize

\caption{Comparison Results on DomainNet under Test-Time Adaptation. UMFC-Memory and UMFC-EMA represent different ways to update the  statics vectors for calibration. }
\vspace{+2mm}
\begin{tabular}{l|cccccc|c}
\toprule
Method   & C& I                 & P                  & Q                 & R   & S & Avg\\
\midrule
CLIP \cite{radford2021clip}    & 71.21  & 49.47  & 64.61  & 14.23  & 82.98  & 64.81  & 57.88   \\
TPT \cite{Shu2022TPT}                 & {73.23}      & {52.63}      & {68.00}      &{12.79}       & {84.39}      & {66.68}      & {59.62}     \\
\cmidrule{1-8}
\textbf{UMFC-Memory}      & {72.82} & {55.12} & {66.82} & 19.92 & 83.62 & 66.82 & {\textbf{60.85}} \\
\textbf{UMFC-EMA}                & {72.99}      & {54.94}      & {66.64}      &  18.58      & 83.54     & 66.75      & {60.57}     \\
\bottomrule
\end{tabular}
\label{DN-on-TTA}
\end{table}

\textbf{Analysis.} \
We mainly compare our UMFC with TPT \cite{Shu2022TPT} in test-time adaptation, where TPT fine-tunes the prompts by minimizing the entropy of the predictions. The comparison results are shown in Table \ref{DN-on-TTA} and Table \ref{IN-on-TTA}. We can observe that: \textbf{(1)} As shown in Table \ref{DN-on-TTA}, UMFC achieves performance gains across all domains, with particularly noticeable gains in those where CLIP performs poorly. 
For example, UMFC substantially improves upon CLIP on ``quickdraw'' and ``infograph'' domains, with an accuracy gain of $5.6\%$.
Table \ref{IN-on-TTA} shows that the improvement on ImageNet-A is less significant than on ImageNet-R and ImageNet-S. 
The reason may be attributed to the absence of distinct domain styles in ImageNet-A,  limiting the effectiveness of our calibration method that relies on domain information.
\textbf{(2)} UMFC-Memory and UMFC-EMA with different statistical update strategies exhibit similar performance. However,  UMFC-EMA, which updates features based on the most recent data, notably reduces storage requirements. Moreover, UMFC exhibits significantly higher computational efficiency without training, whereas TPT requires executing  optimization steps on $64$ different augmented views of each test image. Thus, our method is more suitable for rapid deployment.

\subsection{Ablation Study}

\begin{figure}[t]
    \centering   
    \subfigure[] 
    {
        \begin{minipage}[b]{.45\linewidth} 
            \centering
            \includegraphics[scale=0.22]{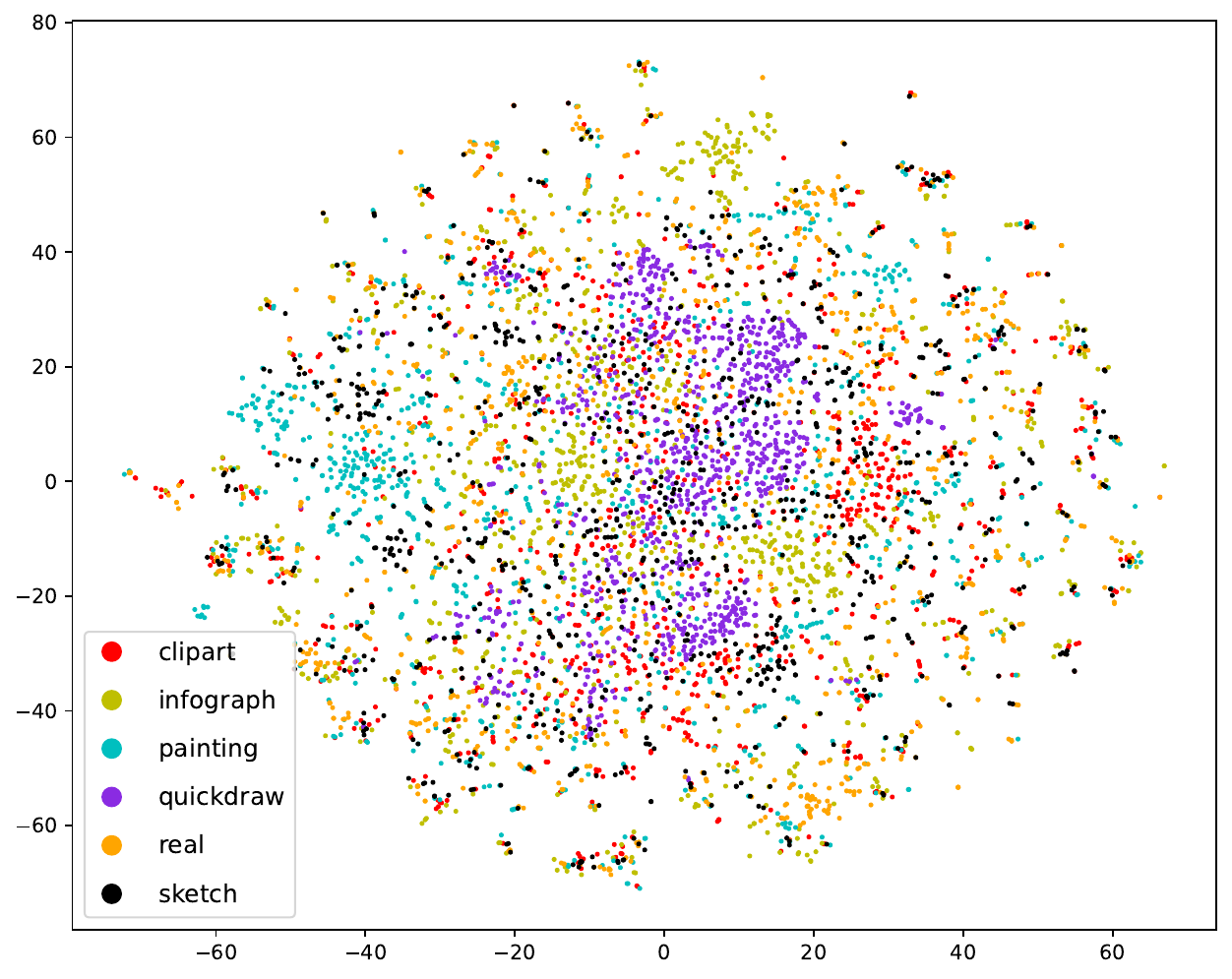}
        \end{minipage}
        \label{image_domain_calibration}
    }
    \subfigure[]
    {
        \begin{minipage}[b]{.45\linewidth}
            \centering
            \includegraphics[scale=0.22]{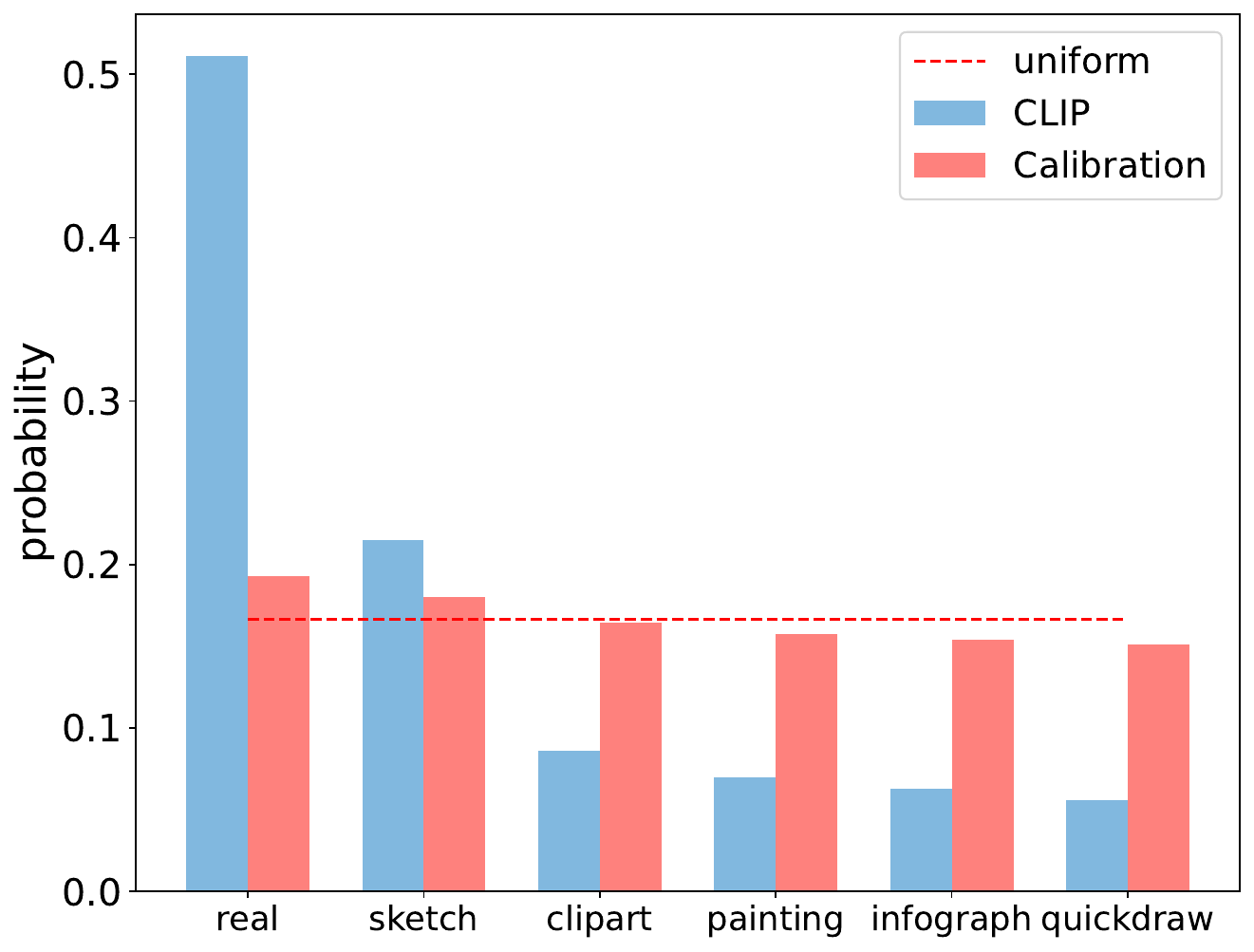}
        \end{minipage}
        \label{text_domain_bias}
    }
    \captionsetup{font={small}}
    \small
    \caption{On DomainNet dataset, we visualize (a) The image features extracted by UMFC image encoder across different domains. (b) The classification probabilities of CLIP's text features on different domains.}
    \label{introduction}
\end{figure}

\textbf{The effectiveness of IFC.} \
We firstly evaluate the effectiveness of IFC. 
As shown in Table \ref{ablation}, IFC individually contributes to performance gains of CLIP, about $3\%$ average gains. 
Furthermore, 
Figure \ref{image_domain_clip} and Figure \ref{image_domain_calibration} visualize the image features extracted by CLIP with/without IFC   respectively.
As shown in Figure \ref{image_domain_clip}, 
the vanilla CLIP maps different-domain images to different clusters in the feature space. 
Conversely, Figure \ref{image_domain_calibration} shows that IFC  leads to the merge of image features from different domains, validating its effectiveness of reducing domain-specific information.

\textbf{The effectiveness of TFC.} \
As shown in Table \ref{ablation}, TFC also individually contributes to performance gains of CLIP. 
To assess the effectiveness of calibrated text features in eliminating domain bias, we construct a domain classifier. Specifically, we utilize the text features generate by CLIP, with domain prompts ``[domain]'', 
as the domain classifer.
Then, we use this domain classifier to perform domain classification for the text features before and after calibration (\ie, computing the cosine similarity between domain classifier and text features). 
As shown in Figure \ref{text_domain_bias}, the original text features exhibit a long-tail phenomenon, with a higher probability of being classified into the ``real'' domain. This suggests that the original text features are biased towards the ``real'' domain, overlooking their generalization ability to other domains.
However, after calibration with TFC, the text features exhibit a near-uniform distribution across domains, indicating that the calibrated text features have largely eliminated domain bias.
Consequently, our method enhances performance in other domains without compromising the model's performance in its strongest domain.
As shown in Table \ref{ablation}, the combination of IFC and TFC (\ie, UMFC) can further bring performance gains, which validates the   complementary of image-level calibration and text-level calibration for CLIP.

\begin{table}[t]
\centering
\captionsetup{font={small}}
\scriptsize

\caption{Ablation study on the effects of TFC and IFC under Transductive Learning.}
\vspace{+2mm}
\begin{tabular}{l|cccccc|c}
\toprule
Method & C     & I     & P     & Q     & R     & S     & Avg   \\
\midrule
CLIP \cite{radford2021clip} & 71.21 & 49.47 & 64.61 & 14.23 & 82.98 & 64.81 & 57.88 \\
\midrule
IFC    & 72.98 & 55.07 & 66.65 & 19.87 & 83.54 & 66.97 & 60.85 \\
TFC    & 71.44 & 49.99 & 65.59 & 13.9  & 83.25 & 64.68 & 58.14 \\
\midrule
\textbf{UMFC}   & 73.01 & 55.44 & 66.89 & 20.14 & 83.66 & 67.51 & \textbf{61.11} \\
\bottomrule
\end{tabular}

\label{ablation}
\end{table}

\paragraph{The impact of cluster number $M$.}
Our method involves clustering the unlabeled data to determine their respective clusters. 
We evaluate our method with respect to the number of clusters $M$. As shown in Table \ref{clusters-k}, our UMFC consistently outperforms vanilla CLIP, even when the number of clusters $M$ does not match the number of domains (6 for DomainNet). More importantly, our method is not sensitive to changes in $M$. 
Refer to the Appendix \ref{appendix:more_analysis} for more analysis.

\begin{table}[t]
\centering
\captionsetup{font={small}}
\small
\scriptsize
\caption{The impact of cluster number $M$ on DomainNet under Transductive Learning.}
\vspace{+2mm}
\begin{tabular}{l|cccccc|c}
\toprule
Method     & C     & I     & P     & Q     & R     & S     & Avg   \\
\midrule
CLIP \cite{radford2021clip} & 71.21 & 49.47 & 64.61 & 14.23 & 82.98 & 64.81 & 57.88 \\
\midrule
UMFC (M=3)  & 72.53 & 54.60 & 66.31 & 20.32 & 83.35 & 66.86 & 60.66 \\
UMFC (M=4)  & 73.55 & 56.36 & 67.19 & 20.62 & 84.13 & 67.69 & 61.59 \\
UMFC (M=6)  & 73.01 & 55.44 & 66.89 & 20.14 & 83.66 & 67.51 & 61.11 \\
UMFC (M=8)  & 73.50 & 56.58 & 67.53 & 20.64 & 84.06 & 67.92 & 61.71 \\
UMFC (M=10) & 73.63 & 56.87 & 67.81 & 20.23 & 84.20 & 67.87 & \textbf{61.77} \\
\bottomrule
\end{tabular}
\label{clusters-k}
\end{table}

\paragraph{The impact of batch size in TTA setting.}
We present results across various batch sizes during test-time adaptation to confirm that UMFC is robust to batch size variations.
When the sample count is initially lower than the number of clusters $M$, K-Means clustering cannot be directly applied. To address this, we used the first $M$ samples as the initial cluster centers and then proceeded with the same test-time adaptation. As shown in Table \ref{TTA_batchsize}, even in the extreme case of a batch size of 1, our method still demonstrates consistent improvement. 

\begin{table}[t]
\centering
\captionsetup{font={small}}
\scriptsize

\caption{The impact of batch size under Test-Time Adaptation.}
\vspace{+2mm}
\begin{tabular}{l|cccccc|c}
\toprule
Method & C     & I     & P     & Q     & R     & S     & Avg   \\
\midrule
CLIP \cite{radford2021clip} & 71.21 & 49.47 & 64.61 & 14.23 & 82.98 & 64.81 & 57.88 \\
\midrule
UMFC (bs=1)   & 72.64 & 53.74 & 66.39 & 18.25 & 83.34 & 66.90 & 60.21  \\
UMFC (bs=10)  & 72.64 & 54.52 & 66.51 & 18.53 & 83.35 & 67.06 & 60.44  \\
UMFC (bs=16)  & 72.70 & 54.80 & 66.91 & 19.11 & 83.78 & 66.69 & 60.66  \\
UMFC (bs=32)  & 73.02 & 55.02 & 66.73 & 19.17 & 83.66 & 66.97 & 60.76  \\
UMFC (bs=64)  & 73.23 & 55.04 & 66.72 & 19.15 & 83.78 & 66.84 & 60.79  \\
UMFC (bs=100) & 72.82 & 55.12 & 66.82 & 19.92 & 83.62 & 66.82 & \textbf{60.85}  \\
\bottomrule
\end{tabular}

\label{TTA_batchsize}
\end{table}

\section{Conclusion}
In this work, we point out that the model biases hinder the transfer ability of pre-trained vision-language models. We develop  UMFC, a simple unsupervised calibration method that mitigates model biases in both visual encoder and text encoder through a training-free manner. We demonstrate the effectiveness of our method across multiple settings, including unsupervised calibration, transductive learning, and test-time adaptation.  Without the need for any annotations and training, UMFC improves the zero-shot generalization ability of CLIP.

\textbf{Limitations and Broader Impacts.} \
Our method requires unlabeled samples from the target domain to establish domain-level calibration vectors. Although our method is much cheaper and more accessible than fine-tuning or few-shot methods \cite{zhou2022coop,zhou2022cocoop,Shu2022TPT}, the requirement of unlabeled data can still be a limitation for some specific scenarios. 

\section*{Acknowledgments}
This work is partially supported by National Key R\&D Program of China no. 2021ZD0111901, National Natural Science Foundation of China (NSFC): 62376259 and 62306301, National Postdoctoral Program for Innovative Talents under Grant BX20220310.

\bibliographystyle{plain}



\newpage

\clearpage

\appendix

\section{The Prevalence of the Observed Model Bias}
\label{appendix:general_observations}
Our method is motivated by observed biases in CLIP. Therefore, it is valuable to investigate whether these observations hold when the pre-training distribution and model architecture are altered.

\begin{figure}[htbp]
    \centerline{\includegraphics[scale=0.4]{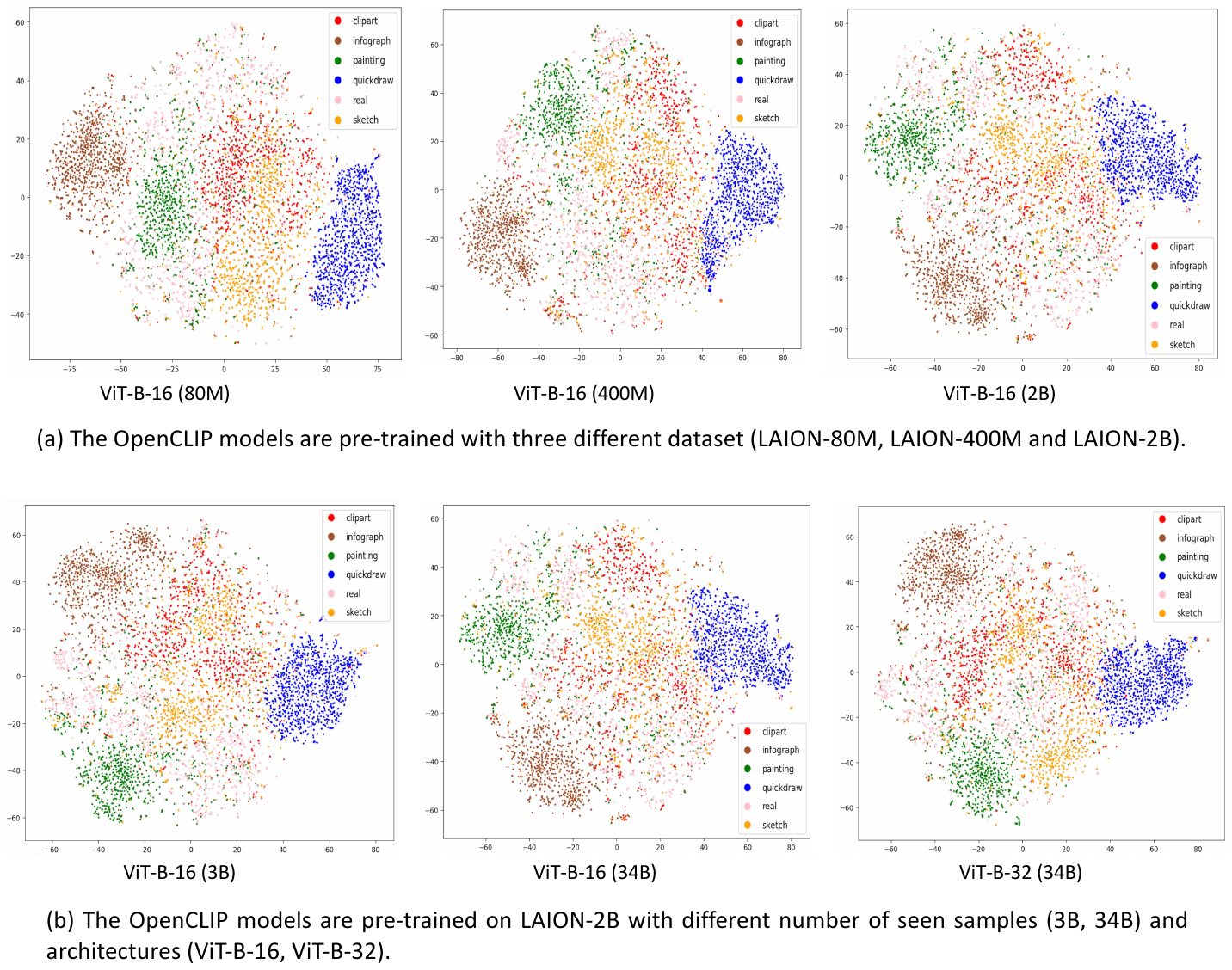}}
    \caption{Visualization of Image Features based on OpenCLIP series.}
    \label{appendix_openclip_visualization}
\end{figure}

We further verify this on OpenCLIP\cite{Cherti2023openclip}. OpenCLIP investigates scaling laws for contrastive language-image pre-training (CLIP) with the public LAION dataset and the open-source OpenCLIP repository. The large-scale experiments involve models trained on up to two billion image-text pairs (LAION-2B) and identify power law scaling for multiple downstream tasks including zero-shot classification, retrieval, linear probing, and end-to-end fine-tuning. We used OpenCLIP models trained on different datasets to obtain the corresponding image features, and visualized them using t-SNE. As shown in Figure \ref{appendix_openclip_visualization}, even when the training data changes, images from different domains still cluster together, demonstrating that our observation is universally applicable.

Additionally, we validated the effectiveness of our method on the OpenCLIP series models in Table \ref{appendix_openclip_umfc}. It is evident that the zero-shot capabilities of larger-scale models have significantly improved compared to CLIP, and our method still provides further performance gains. This demonstrates that our method is universally applicable to Vision-Language models with different architectures and training data.

\begin{table}[H]
\centering
\captionsetup{font={small}}
\small
\scriptsize
\caption{Comparision Results on DominaNet using OpenCLIP.}
\vspace{+2mm}
\begin{tabular}{l|cccccc|c}
\toprule
Arch     & C     & I     & P     & Q     & R     & S     & Avg   \\
\midrule
ViT-B-16 & 78.66 & 56.16 & 71.09 & 15.25 & 86.93 & 72.26 & 63.39 \\
+ UMFC   & 78.93 & 58.29 & 71.78 & 21.56 & 86.36 & 73.29 & \textbf{65.04} \\
\midrule
ViT-B-32 & 76.98 & 52.00 & 68.77 & 15.63 & 85.63 & 70.54 & 61.59 \\
+ UMFC   & 76.99 & 53.08 & 68.70 & 22.61 & 85.17 & 71.27 & \textbf{62.97} \\
\midrule
ViT-H-14 & 83.72 & 63.21 & 76.05 & 18.20 & 89.49 & 79.18 & 68.31 \\
+ UMFC   & 83.74 & 63.89 & 76.41 & 23.73 & 89.33 & 79.49 & \textbf{69.43} \\
\bottomrule
\end{tabular}
\label{appendix_openclip_umfc}
\end{table}

\section{Details of Text Feature Calibration Module}
\label{appendix: TFC details}
\begin{figure}[htbp]
    \centerline{\includegraphics[scale=0.2]{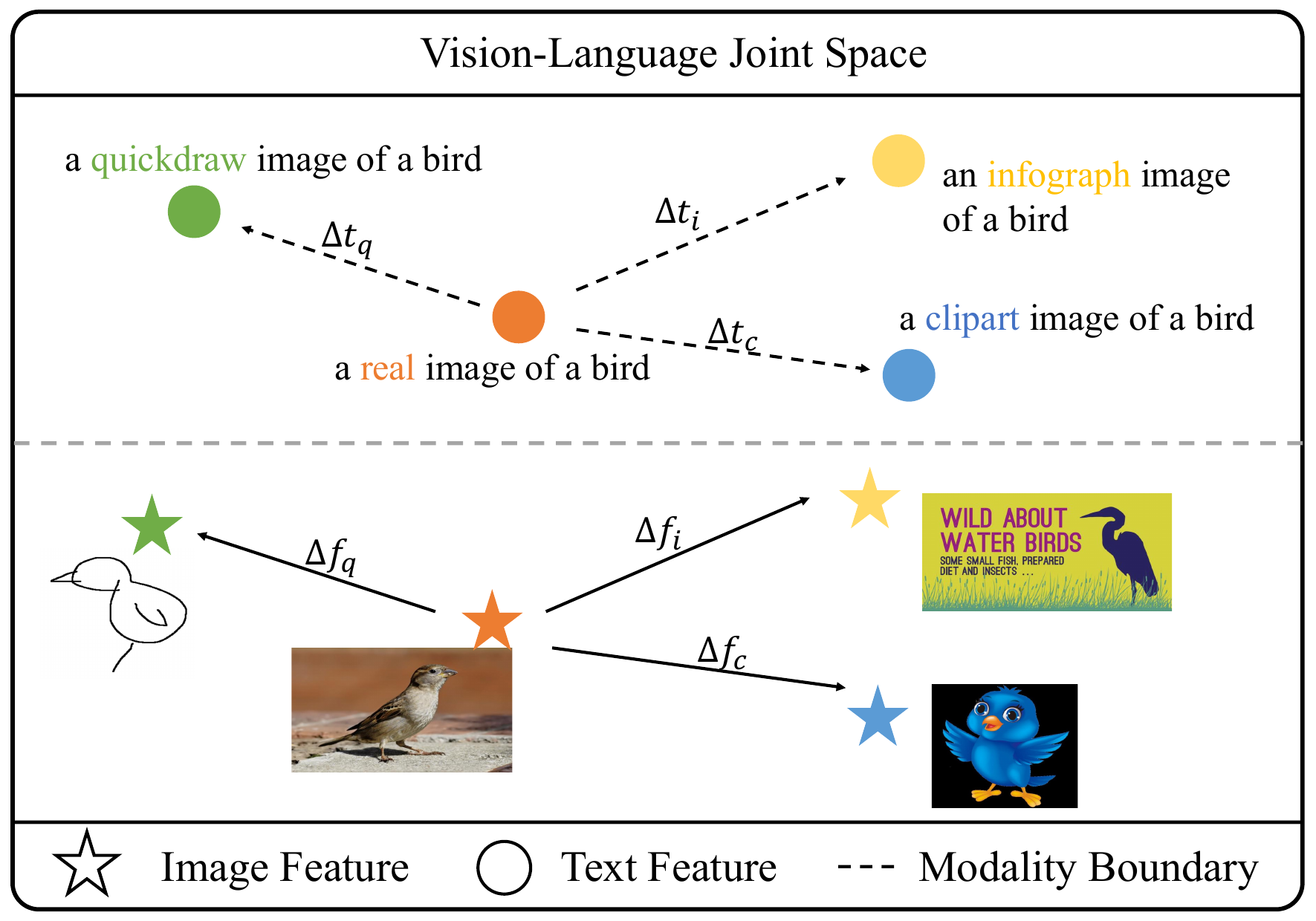}}
    \caption{The domain transition direction between texts is similar to that between images. }
    \label{appendix_VL-space}
\end{figure}

Since the domain labels are unavailable, we attempt to extract domain information from images in downstream tasks, and then transfer this domain information to the text embedding space to eliminate the text bias. 
Inspired by Observation \ref{global_direction}, we can estimate the domain transition direction in text embedding space by using images from different domains, as shown in  Figure \ref{appendix_VL-space}. 

\section{Algorithm}
\label{appendix: algorithm}

Algorithm \ref{alg_TTA} summarizes the proposed UMFC method under Test-Time Adaptation (TTA) setting. 

Algorithm \ref{alg_TL} summarizes the proposed UMFC method under Unsupervised Calibration (UC) / Transductive Learning (TL). 

\begin{algorithm}[H]
	\caption{Algorithm UMFC under TTA}
	\label{alg_TTA}
	\begin{algorithmic}[1]
    \State {\bfseries Input}: test data $\left\{u_{b,i}\right\}_{b=1}^{B}{}_{i=1}^{N}$, batch size $B$, total number of batches $N$, total number of clusters $M$, CLIP's image encoder $F(\cdot)$ and text encoder $T(p;\cdot)$, $p$ is the text template, class names $\{y_i\}_{i=1}^C$.
        \State $t \leftarrow T(p;\{y_i\}_{i=1}^C)$ \Comment{Get original text features}
        \State Initialize centroids of $M$ cluster $\{c_j\}_{j=1}^M$ as empty set
    \For{$i\leftarrow 1$  {\bfseries to} $N$} 
        \State $\{f_{b,i}\}_{b=1}^B \leftarrow F(\{u_{b,i}\}_{b=1}^B)$ \Comment{Get original image features of $i$-th batch}
        \If{$\{c_j\}_{j=1}^M$ is $\emptyset$} 
        \State $\{c_j\}_{j=1}^M \leftarrow$ K-Means$(\{f_{b,i}\}_{b=1}^B)$  \Comment{Using K-Means to acquire $M$ centroids}
        \EndIf

    \For {$b\leftarrow 1$  {\bfseries to} $B$} 
        \State $l_{b,i} \leftarrow \mathop{\arg\min}\limits_{m}\|f_{b,i} - c_m\|_2$ \Comment{Assign sample to the nearest clusters }
    \EndFor
    
    \State $c_k' \leftarrow \frac{1}{B}\sum_{b=1}^B\mathbb{I}(l_{b,i}=m)\cdot f_{b,i}$ \Comment{Calculate centroids of current batch}
    \State $c_k \leftarrow c_k + \eta \cdot c_k'$, \Comment{Update prototypes, where $\eta$ is the update weight}
    \State $\mu_{avg} \leftarrow \frac{1}{M} \sum_{m=1}^{M}c'_m$ \Comment{update the average image feature}
        
    \State  $\hat{t}^m \leftarrow \mathbb{I}(l=m)(c'_m - \mu_{avg})$, \Comment{Update calibration statics}
    \State $t'_{b} \leftarrow \frac{1}{M}\sum_{m=1}^{M}{\frac{{t - \hat{t}^m}}{\|{t - \hat{t}^m}\|}}$, 
    \For {$b\leftarrow 1$  {\bfseries to} $B$}
    \State $f'_{b,i} \leftarrow \frac{f_{b,i} - c'_l}{\|f_{b,i} - c'_l\|}$, \Comment{Calibrate $f$ and $t$  with IFC and TFC}
    \EndFor
        
    \EndFor
    \State \Return $\{f'_{b,i}, t'_{b}\}_{b=1}^B{}_{i=1}^N $ \Comment{Calibrated image and text features}
	\end{algorithmic}  
\end{algorithm}

\begin{algorithm}[H]
	\caption{Algorithm UMFC under UC / TL}
	\label{alg_TL}
	\begin{algorithmic}[1]
		\State {\bfseries Input}: unlabeled data $\left\{u_b\right\}_{i=1}^{N}$, total number of samples $N$, total number of clusters $M$, total number of batches $B$, CLIP's image encoder $F(\cdot)$ and text encoder $T(p;\cdot)$, $p$ is the text template, class names $\{y_i\}_{i=1}^C$.
        \State $t \leftarrow T(p;\{y_i\}_{i=1}^C)$ \Comment{Get original text features}
        \State $\{f_b\}_{b=1}^N \leftarrow F(\{u_b\}_{i=1}^{N})$ \Comment{Get original image features of unlabeled data}
        \State $\mu_{avg} \leftarrow \frac{1}{N} \sum_{b=1}^{N}f_b$ \Comment{Get the average of $\{f_b\}_{b=1}^N $}
        \State $\{c_j\}_{j=1}^M \leftarrow$ K-Means$(\{f_b\}_{b=1}^N)$  \Comment{Using K-Means to acquire $M$ clusters and their centroids}
    \For{$b\leftarrow 1$  {\bfseries to} $N$} 

        \State $l \leftarrow \mathop{\arg\min}\limits_{m}\|f_b - c_m\|_2$ \Comment{Assign sample to the nearest clusters }
        \State  $\hat{t}_k \leftarrow \mathbb{I}(l=m)(c_m - \mu_{avg})$, \Comment{Update calibration statics}
        \State $f'_b \leftarrow \frac{f_b - c_l}{\|f_b - c_l\|}$, $t'_b \leftarrow \frac{1}{M}\sum_{m=1}^{M}{\frac{{t - \hat{t}^m}}{\|{t - \hat{t}^m}\|}}$, \Comment{Calibrate $f$ and $t$  with IFC and TFC}
    \EndFor
    \State \Return $\{f'_b\}_{b=1}^N $, $t' $ \Comment{Calibrated image and text features}
	\end{algorithmic}  
\end{algorithm}

\section{Experimental Settings}
\label{appendix:experimental-setting}
Our work can be deployed across various scenarios, including Unsupervised Calibration (UC), Test-Time Adaptation (TTA), and Transductive Learning (TL).

\textbf{Unsupervised Calibration.} In the UC scenario, we provide an unsupervised training set and use K-Means to assign cluster labels to the training samples, while also saving the corresponding cluster prototypes. Then, we create an unlabeled training set from a mixed domain by sampling 16 instances from each class across 6 domains in DomainNet. UMFC derives image bias and text bias for different clusters based on the cluster labels. During the testing phase, we first predict the cluster labels of the test samples using the cluster prototypes obtained during training, and then calibrate the predictions using the bias information derived from UMFC.   

\textbf{Test-Time Adaptation.} In the TTA scenario, no training data is provided. Test data from mixed domains arrive in batches, with a batch size set to 100. For the first batch, we perform initial clustering using K-Means. For subsequent batches, we assign cluster labels to the samples based on cluster prototypes and continuously update these prototypes. Once the cluster labels are obtained, UMFC calculates the bias information for the current batch, updates the bias information for each cluster based on the new labels, and then calibrates the data for the current batch. 

\textbf{Transductive Learning.} In the TL scenario, we provide the entire test set. Similar to UMFC, we gather statistical information and calibrate the predictions for the test data. TL can be viewed as an extreme case of TTA, where the entire test set is treated as a single batch.

\section{Experimental Analysis}
\label{appendix:more_analysis}
\paragraph{The impact of cluster number $M$.}
We evaluate our method with respect to the number of clusters $M$ and demonstrate that our method is not sensitive to the choice of this hyperparameter. As shown in Table \ref{IN_clusters-k}, For instance, setting $M$ from 2 to 6 all leads to improvements on ImageNet-Variants.  

\begin{table}[H]
\centering
\captionsetup{font={small}}
\small
\scriptsize
\caption{The impact of cluster number $M$ on ImageNet-Variants under Transductive Learning.}
\vspace{+2mm}
\begin{tabular}{l|ccc|c}
\toprule
Method     & IN-A  & IN-R  & IN-S  & Avg    \\
\midrule
CLIP \cite{radford2021clip}       & 42.13 & 66.95 & 74.58 & 61.22 \\
UMFC (M=2)  & 45.35 & 71.71 & 77.37 & 64.81 \\
UMFC (M=3)  & 44.77 & 72.19 & 78.62 & \textbf{65.19} \\
UMFC (M=6)  & 45.29 & 71.33 & 77.59 & 64.74 \\
\bottomrule
\end{tabular}
\label{IN_clusters-k}
\end{table}

\paragraph{Computation Cost.}
For analyzing computational cost, we report the training time, inference time and memory of UMFC and other comparison methods under different scenarios.
In \textbf{Unsupervised Calibration} scenario, the entire unlabeled training set is provided for training. Corresponding computation cost comparisons are shown in Table \ref{compute_cost_TL}. Firstly, UMFC incurs minimal training and inference overhead compared to CLIP. This is because UMFC only requires a single forward pass to extract features and then calculate statistics for feature calibration. Secondly, when compared to few-shot fine-tuning methods (CoOp) and unsupervised fine-tuning methods (MUST), UMFC also demonstrates lower consumption of computational resources and time.
In \textbf{Test-time Adaptation} scenario, no training data is provided and the test data arrives in batches. Corresponding computation cost comparisons are shown in Table \ref{compute_cost_TTA}. UMFC requires less memory than TPT and shows greater computational efficiency. Specifically, UMFC takes only 296 seconds, whereas TPT requires nearly 197 minutes. This is because TPT requires fine-tuning the text prompt for each test sample and augmenting each test sample 64 times to ensure the reliability of the fine-tuning results, which significantly slows down TPT's inference speed.

\begin{table}[H]
\centering
\captionsetup{font={small}}
\small
\scriptsize
\caption{Computation Cost under Transductive Learning.}
\vspace{+2mm}
\begin{tabular}{l|cccc}
\toprule
Method           & Training Time            & Inference Time & Epoch & Memory   \\
\midrule
CLIP \cite{radford2021clip} & -                        & 86 seconds     & -     & 1797MiB  \\
MUST \cite{Li2022MUST}            & 10 hours (2 GPUs)        & 92 seconds     & 30    & 25944MiB \\
\midrule
CoOp (6*1 shot) \cite{zhou2022coop} & 32 minutes               & 83 seconds     & 50    & 7007MiB  \\
CoOp (6*4 shots) \cite{zhou2022coop}& 160 minutes              & 83 seconds     & 100   & 7007MiB  \\
\midrule
UMFC             & 57.3 seconds   & 86 seconds     & -     & 1887MiB  \\
\bottomrule
\end{tabular}
\label{compute_cost_TL}
\end{table}

\begin{table}[H]
\centering
\captionsetup{font={small}}
\small
\scriptsize
\caption{Computation Cost under Test-Time Adaptation.}
\vspace{+2mm}
\begin{tabular}{l|cc}
\toprule
Method  & Inference Time           & Memory     \\
\midrule
TPT  & 197 minutes              & 6872MiB    \\
UMFC & 296 seconds              & 1790MiB    \\
\bottomrule
\end{tabular}
\label{compute_cost_TTA}
\end{table}


\newpage
\clearpage

\section*{NeurIPS Paper Checklist}

\begin{enumerate}

\item {\bf Claims}
    \item[] Question: Do the main claims made in the abstract and introduction accurately reflect the paper's contributions and scope?
    \item[] Answer: \answerYes{} 
    \item[] Justification: A summary of the paper's contributions is provided at conclusion.
    \item[] Guidelines:
    \begin{itemize}
        \item The answer NA means that the abstract and introduction do not include the claims made in the paper.
        \item The abstract and/or introduction should clearly state the claims made, including the contributions made in the paper and important assumptions and limitations. A No or NA answer to this question will not be perceived well by the reviewers. 
        \item The claims made should match theoretical and experimental results, and reflect how much the results can be expected to generalize to other settings. 
        \item It is fine to include aspirational goals as motivation as long as it is clear that these goals are not attained by the paper. 
    \end{itemize}

\item {\bf Limitations}
    \item[] Question: Does the paper discuss the limitations of the work performed by the authors?
    \item[] Answer: \answerYes{} 
    \item[] Justification: See Section 6.
    \item[] Guidelines:
    \begin{itemize}
        \item The answer NA means that the paper has no limitation while the answer No means that the paper has limitations, but those are not discussed in the paper. 
        \item The authors are encouraged to create a separate "Limitations" section in their paper.
        \item The paper should point out any strong assumptions and how robust the results are to violations of these assumptions (e.g., independence assumptions, noiseless settings, model well-specification, asymptotic approximations only holding locally). The authors should reflect on how these assumptions might be violated in practice and what the implications would be.
        \item The authors should reflect on the scope of the claims made, e.g., if the approach was only tested on a few datasets or with a few runs. In general, empirical results often depend on implicit assumptions, which should be articulated.
        \item The authors should reflect on the factors that influence the performance of the approach. For example, a facial recognition algorithm may perform poorly when image resolution is low or images are taken in low lighting. Or a speech-to-text system might not be used reliably to provide closed captions for online lectures because it fails to handle technical jargon.
        \item The authors should discuss the computational efficiency of the proposed algorithms and how they scale with dataset size.
        \item If applicable, the authors should discuss possible limitations of their approach to address problems of privacy and fairness.
        \item While the authors might fear that complete honesty about limitations might be used by reviewers as grounds for rejection, a worse outcome might be that reviewers discover limitations that aren't acknowledged in the paper. The authors should use their best judgment and recognize that individual actions in favor of transparency play an important role in developing norms that preserve the integrity of the community. Reviewers will be specifically instructed to not penalize honesty concerning limitations.
    \end{itemize}

\item {\bf Theory Assumptions and Proofs}
    \item[] Question: For each theoretical result, does the paper provide the full set of assumptions and a complete (and correct) proof?
    \item[] Answer: \answerNA{} 
    \item[] Justification: Not Applicable.
    \item[] Guidelines:
    \begin{itemize}
        \item The answer NA means that the paper does not include theoretical results. 
        \item All the theorems, formulas, and proofs in the paper should be numbered and cross-referenced.
        \item All assumptions should be clearly stated or referenced in the statement of any theorems.
        \item The proofs can either appear in the main paper or the supplemental material, but if they appear in the supplemental material, the authors are encouraged to provide a short proof sketch to provide intuition. 
        \item Inversely, any informal proof provided in the core of the paper should be complemented by formal proofs provided in appendix or supplemental material.
        \item Theorems and Lemmas that the proof relies upon should be properly referenced. 
    \end{itemize}

    \item {\bf Experimental Result Reproducibility}
    \item[] Question: Does the paper fully disclose all the information needed to reproduce the main experimental results of the paper to the extent that it affects the main claims and/or conclusions of the paper (regardless of whether the code and data are provided or not)?
    \item[] Answer: \answerYes{} 
    \item[] Justification: See Section 5.2, 5.4 and supplemental material for implementation details.
    \item[] Guidelines:
    \begin{itemize}
        \item The answer NA means that the paper does not include experiments.
        \item If the paper includes experiments, a No answer to this question will not be perceived well by the reviewers: Making the paper reproducible is important, regardless of whether the code and data are provided or not.
        \item If the contribution is a dataset and/or model, the authors should describe the steps taken to make their results reproducible or verifiable. 
        \item Depending on the contribution, reproducibility can be accomplished in various ways. For example, if the contribution is a novel architecture, describing the architecture fully might suffice, or if the contribution is a specific model and empirical evaluation, it may be necessary to either make it possible for others to replicate the model with the same dataset, or provide access to the model. In general. releasing code and data is often one good way to accomplish this, but reproducibility can also be provided via detailed instructions for how to replicate the results, access to a hosted model (e.g., in the case of a large language model), releasing of a model checkpoint, or other means that are appropriate to the research performed.
        \item While NeurIPS does not require releasing code, the conference does require all submissions to provide some reasonable avenue for reproducibility, which may depend on the nature of the contribution. For example
        \begin{enumerate}
            \item If the contribution is primarily a new algorithm, the paper should make it clear how to reproduce that algorithm.
            \item If the contribution is primarily a new model architecture, the paper should describe the architecture clearly and fully.
            \item If the contribution is a new model (e.g., a large language model), then there should either be a way to access this model for reproducing the results or a way to reproduce the model (e.g., with an open-source dataset or instructions for how to construct the dataset).
            \item We recognize that reproducibility may be tricky in some cases, in which case authors are welcome to describe the particular way they provide for reproducibility. In the case of closed-source models, it may be that access to the model is limited in some way (e.g., to registered users), but it should be possible for other researchers to have some path to reproducing or verifying the results.
        \end{enumerate}
    \end{itemize}

\item {\bf Open access to data and code}
    \item[] Question: Does the paper provide open access to the data and code, with sufficient instructions to faithfully reproduce the main experimental results, as described in supplemental material?
    \item[] Answer: \answerYes{} 
    \item[] Justification: See \url{https://github.com/GIT-LJc/UMFC}
    \item[] Guidelines:
    \begin{itemize}
        \item The answer NA means that paper does not include experiments requiring code.
        \item Please see the NeurIPS code and data submission guidelines (\url{https://nips.cc/public/guides/CodeSubmissionPolicy}) for more details.
        \item While we encourage the release of code and data, we understand that this might not be possible, so “No” is an acceptable answer. Papers cannot be rejected simply for not including code, unless this is central to the contribution (e.g., for a new open-source benchmark).
        \item The instructions should contain the exact command and environment needed to run to reproduce the results. See the NeurIPS code and data submission guidelines (\url{https://nips.cc/public/guides/CodeSubmissionPolicy}) for more details.
        \item The authors should provide instructions on data access and preparation, including how to access the raw data, preprocessed data, intermediate data, and generated data, etc.
        \item The authors should provide scripts to reproduce all experimental results for the new proposed method and baselines. If only a subset of experiments are reproducible, they should state which ones are omitted from the script and why.
        \item At submission time, to preserve anonymity, the authors should release anonymized versions (if applicable).
        \item Providing as much information as possible in supplemental material (appended to the paper) is recommended, but including URLs to data and code is permitted.
    \end{itemize}

\item {\bf Experimental Setting/Details}
    \item[] Question: Does the paper specify all the training and test details (e.g., data splits, hyperparameters, how they were chosen, type of optimizer, etc.) necessary to understand the results?
    \item[] Answer: \answerYes{} 
    \item[] Justification: See Section 5.2, 5.4 and supplemental material for implementation details. 
    \item[] Guidelines:
    \begin{itemize}
        \item The answer NA means that the paper does not include experiments.
        \item The experimental setting should be presented in the core of the paper to a level of detail that is necessary to appreciate the results and make sense of them.
        \item The full details can be provided either with the code, in appendix, or as supplemental material.
    \end{itemize}

\item {\bf Experiment Statistical Significance}
    \item[] Question: Does the paper report error bars suitably and correctly defined or other appropriate information about the statistical significance of the experiments?
    \item[] Answer: \answerNA{} 
    \item[] Justification: Not Applicable.
    \item[] Guidelines:
    \begin{itemize}
        \item The answer NA means that the paper does not include experiments.
        \item The authors should answer "Yes" if the results are accompanied by error bars, confidence intervals, or statistical significance tests, at least for the experiments that support the main claims of the paper.
        \item The factors of variability that the error bars are capturing should be clearly stated (for example, train/test split, initialization, random drawing of some parameter, or overall run with given experimental conditions).
        \item The method for calculating the error bars should be explained (closed form formula, call to a library function, bootstrap, etc.)
        \item The assumptions made should be given (e.g., Normally distributed errors).
        \item It should be clear whether the error bar is the standard deviation or the standard error of the mean.
        \item It is OK to report 1-sigma error bars, but one should state it. The authors should preferably report a 2-sigma error bar than state that they have a 96\% CI, if the hypothesis of Normality of errors is not verified.
        \item For asymmetric distributions, the authors should be careful not to show in tables or figures symmetric error bars that would yield results that are out of range (e.g. negative error rates).
        \item If error bars are reported in tables or plots, The authors should explain in the text how they were calculated and reference the corresponding figures or tables in the text.
    \end{itemize}

\item {\bf Experiments Compute Resources}
    \item[] Question: For each experiment, does the paper provide sufficient information on the computer resources (type of compute workers, memory, time of execution) needed to reproduce the experiments?
    \item[] Answer: \answerYes{} 
    \item[] Justification: See Section 5.2 and supplemental material for implementation details.
    \item[] Guidelines:
    \begin{itemize}
        \item The answer NA means that the paper does not include experiments.
        \item The paper should indicate the type of compute workers CPU or GPU, internal cluster, or cloud provider, including relevant memory and storage.
        \item The paper should provide the amount of compute required for each of the individual experimental runs as well as estimate the total compute. 
        \item The paper should disclose whether the full research project required more compute than the experiments reported in the paper (e.g., preliminary or failed experiments that didn't make it into the paper). 
    \end{itemize}
    
\item {\bf Code Of Ethics}
    \item[] Question: Does the research conducted in the paper conform, in every respect, with the NeurIPS Code of Ethics \url{https://neurips.cc/public/EthicsGuidelines}?
    \item[] Answer: \answerYes{} 
    \item[] Justification: The research conducted in the paper conform, in every respect, with the NeurIPS Code of Ethics
    \item[] Guidelines:
    \begin{itemize}
        \item The answer NA means that the authors have not reviewed the NeurIPS Code of Ethics.
        \item If the authors answer No, they should explain the special circumstances that require a deviation from the Code of Ethics.
        \item The authors should make sure to preserve anonymity (e.g., if there is a special consideration due to laws or regulations in their jurisdiction).
    \end{itemize}

\item {\bf Broader Impacts}
    \item[] Question: Does the paper discuss both potential positive societal impacts and negative societal impacts of the work performed?
    \item[] Answer: \answerYes{} 
    \item[] Justification: See Section 6.
    \item[] Guidelines:
    \begin{itemize}
        \item The answer NA means that there is no societal impact of the work performed.
        \item If the authors answer NA or No, they should explain why their work has no societal impact or why the paper does not address societal impact.
        \item Examples of negative societal impacts include potential malicious or unintended uses (e.g., disinformation, generating fake profiles, surveillance), fairness considerations (e.g., deployment of technologies that could make decisions that unfairly impact specific groups), privacy considerations, and security considerations.
        \item The conference expects that many papers will be foundational research and not tied to particular applications, let alone deployments. However, if there is a direct path to any negative applications, the authors should point it out. For example, it is legitimate to point out that an improvement in the quality of generative models could be used to generate deepfakes for disinformation. On the other hand, it is not needed to point out that a generic algorithm for optimizing neural networks could enable people to train models that generate Deepfakes faster.
        \item The authors should consider possible harms that could arise when the technology is being used as intended and functioning correctly, harms that could arise when the technology is being used as intended but gives incorrect results, and harms following from (intentional or unintentional) misuse of the technology.
        \item If there are negative societal impacts, the authors could also discuss possible mitigation strategies (e.g., gated release of models, providing defenses in addition to attacks, mechanisms for monitoring misuse, mechanisms to monitor how a system learns from feedback over time, improving the efficiency and accessibility of ML).
    \end{itemize}
    
\item {\bf Safeguards}
    \item[] Question: Does the paper describe safeguards that have been put in place for responsible release of data or models that have a high risk for misuse (e.g., pretrained language models, image generators, or scraped datasets)?
    \item[] Answer: \answerNA{} 
    \item[] Justification: Not Applicable.
    \item[] Guidelines:
    \begin{itemize}
        \item The answer NA means that the paper poses no such risks.
        \item Released models that have a high risk for misuse or dual-use should be released with necessary safeguards to allow for controlled use of the model, for example by requiring that users adhere to usage guidelines or restrictions to access the model or implementing safety filters. 
        \item Datasets that have been scraped from the Internet could pose safety risks. The authors should describe how they avoided releasing unsafe images.
        \item We recognize that providing effective safeguards is challenging, and many papers do not require this, but we encourage authors to take this into account and make a best faith effort.
    \end{itemize}

\item {\bf Licenses for existing assets}
    \item[] Question: Are the creators or original owners of assets (e.g., code, data, models), used in the paper, properly credited and are the license and terms of use explicitly mentioned and properly respected?
    \item[] Answer: \answerYes{} 
    \item[] Justification: See References.
    \item[] Guidelines:
    \begin{itemize}
        \item The answer NA means that the paper does not use existing assets.
        \item The authors should cite the original paper that produced the code package or dataset.
        \item The authors should state which version of the asset is used and, if possible, include a URL.
        \item The name of the license (e.g., CC-BY 4.0) should be included for each asset.
        \item For scraped data from a particular source (e.g., website), the copyright and terms of service of that source should be provided.
        \item If assets are released, the license, copyright information, and terms of use in the package should be provided. For popular datasets, \url{paperswithcode.com/datasets} has curated licenses for some datasets. Their licensing guide can help determine the license of a dataset.
        \item For existing datasets that are re-packaged, both the original license and the license of the derived asset (if it has changed) should be provided.
        \item If this information is not available online, the authors are encouraged to reach out to the asset's creators.
    \end{itemize}

\item {\bf New Assets}
    \item[] Question: Are new assets introduced in the paper well documented and is the documentation provided alongside the assets?
    \item[] Answer: \answerNA{} 
    \item[] Justification: Not Applicable.
    \item[] Guidelines:
    \begin{itemize}
        \item The answer NA means that the paper does not release new assets.
        \item Researchers should communicate the details of the dataset/code/model as part of their submissions via structured templates. This includes details about training, license, limitations, etc. 
        \item The paper should discuss whether and how consent was obtained from people whose asset is used.
        \item At submission time, remember to anonymize your assets (if applicable). You can either create an anonymized URL or include an anonymized zip file.
    \end{itemize}

\item {\bf Crowdsourcing and Research with Human Subjects}
    \item[] Question: For crowdsourcing experiments and research with human subjects, does the paper include the full text of instructions given to participants and screenshots, if applicable, as well as details about compensation (if any)? 
    \item[] Answer: \answerNA{} 
    \item[] Justification: Not Applicable.
    \item[] Guidelines:
    \begin{itemize}
        \item The answer NA means that the paper does not involve crowdsourcing nor research with human subjects.
        \item Including this information in the supplemental material is fine, but if the main contribution of the paper involves human subjects, then as much detail as possible should be included in the main paper. 
        \item According to the NeurIPS Code of Ethics, workers involved in data collection, curation, or other labor should be paid at least the minimum wage in the country of the data collector. 
    \end{itemize}

\item {\bf Institutional Review Board (IRB) Approvals or Equivalent for Research with Human Subjects}
    \item[] Question: Does the paper describe potential risks incurred by study participants, whether such risks were disclosed to the subjects, and whether Institutional Review Board (IRB) approvals (or an equivalent approval/review based on the requirements of your country or institution) were obtained?
    \item[] Answer: \answerNA{} 
    \item[] Justification: Not Applicable.
    \item[] Guidelines:
    \begin{itemize}
        \item The answer NA means that the paper does not involve crowdsourcing nor research with human subjects.
        \item Depending on the country in which research is conducted, IRB approval (or equivalent) may be required for any human subjects research. If you obtained IRB approval, you should clearly state this in the paper. 
        \item We recognize that the procedures for this may vary significantly between institutions and locations, and we expect authors to adhere to the NeurIPS Code of Ethics and the guidelines for their institution. 
        \item For initial submissions, do not include any information that would break anonymity (if applicable), such as the institution conducting the review.
    \end{itemize}

\end{enumerate}


\end{document}